\documentclass[pmlr,twocolumn,10pt]{jmlr} %

\usepackage[utf8]{inputenc}

\usepackage{graphicx}
\usepackage{booktabs}
\usepackage{mathtools}
\usepackage{setspace}
\usepackage{placeins}
\usepackage{tikz}
\usepackage{amsmath}
\usepackage{multirow}
\usepackage{pgfplots}
\pgfplotsset{compat=1.7}
\usepackage{tikz-network}
\usetikzlibrary{patterns}

\usepgfplotslibrary{groupplots}
\usepgfplotslibrary{statistics}

\usepackage{xcolor,pifont}

\usepackage{colortbl}

\definecolor{amber}{rgb}{1.0, 0.75, 0.0}
\definecolor{awesome}{rgb}{1.0, 0.13, 0.32}
\definecolor{ao(english)}{rgb}{0.0, 0.5, 0.0}

\title[]{Explainable Biomedical Recommendations via Reinforcement Learning Reasoning on Knowledge Graphs}

 \author{%
\Name{Gavin Edwards$^1$} \Email{gavin.edwards@astrazeneca.com}
\AND
\Name{Sebastian Nilsson$^2$} \Email{sebastian.nilsson@astrazeneca.com}
\AND
\Name{Benedek Rozemberczki$^1$} \Email{benedek.rozemberczki@astrazeneca.com}
\AND
\Name{Eliseo Papa$^1$} \Email{eliseo.papa@astrazeneca.com}
\AND
\addr $^1$Biological Insight Knowledge Graph (BIKG), Research D\&A, R\&D IT, AstraZeneca, Cambridge, UK
\AND
\addr $^2$Biological Insight Knowledge Graph (BIKG), Research D\&A, R\&D IT, AstraZeneca, Gothenburg, Sweden
}

\begin{document}

\maketitle
\begin{abstract}

For Artificial Intelligence to have a greater impact in biology and medicine, it is crucial that recommendations are both accurate and transparent. In other domains, a neurosymbolic approach of multi-hop reasoning on knowledge graphs has been shown to produce transparent explanations. However, there is a lack of research applying it to complex biomedical datasets and problems. In this paper, the approach is explored for drug discovery to draw solid conclusions on its applicability. For the first time, we systematically apply it to multiple biomedical datasets and recommendation tasks with fair benchmark comparisons. The approach is found to outperform the best baselines by 21.7\% on average whilst producing novel, biologically relevant explanations.
\end{abstract}

\section{Introduction}

Drug development costs are ten times what the industry spent per year in the 1980s and continously growing. The average new drug is now expected to take ten or more years to develop and cost over \$2 billion \citep{austin_hayford_2021}. To improve on this trend, there is increasing interest in enhancing human decision making with Artificial Intelligence (AI) recommendation systems. 

AI recommendation systems commonly use black-box models that solely focus on optimising performance for tasks such as product recommendation and personalised content. These are high-frequency, low-risk recommendations that receive fast feedback to improve and build trust. However, the application of recommendations in the drug discovery space suffers from significant limitations \citep{gogleva2021knowledge}: decisions are infrequent, high-risk and can have a large impact. 

\begin{figure}[t!]
    \centering 
    \includegraphics[scale=0.17]{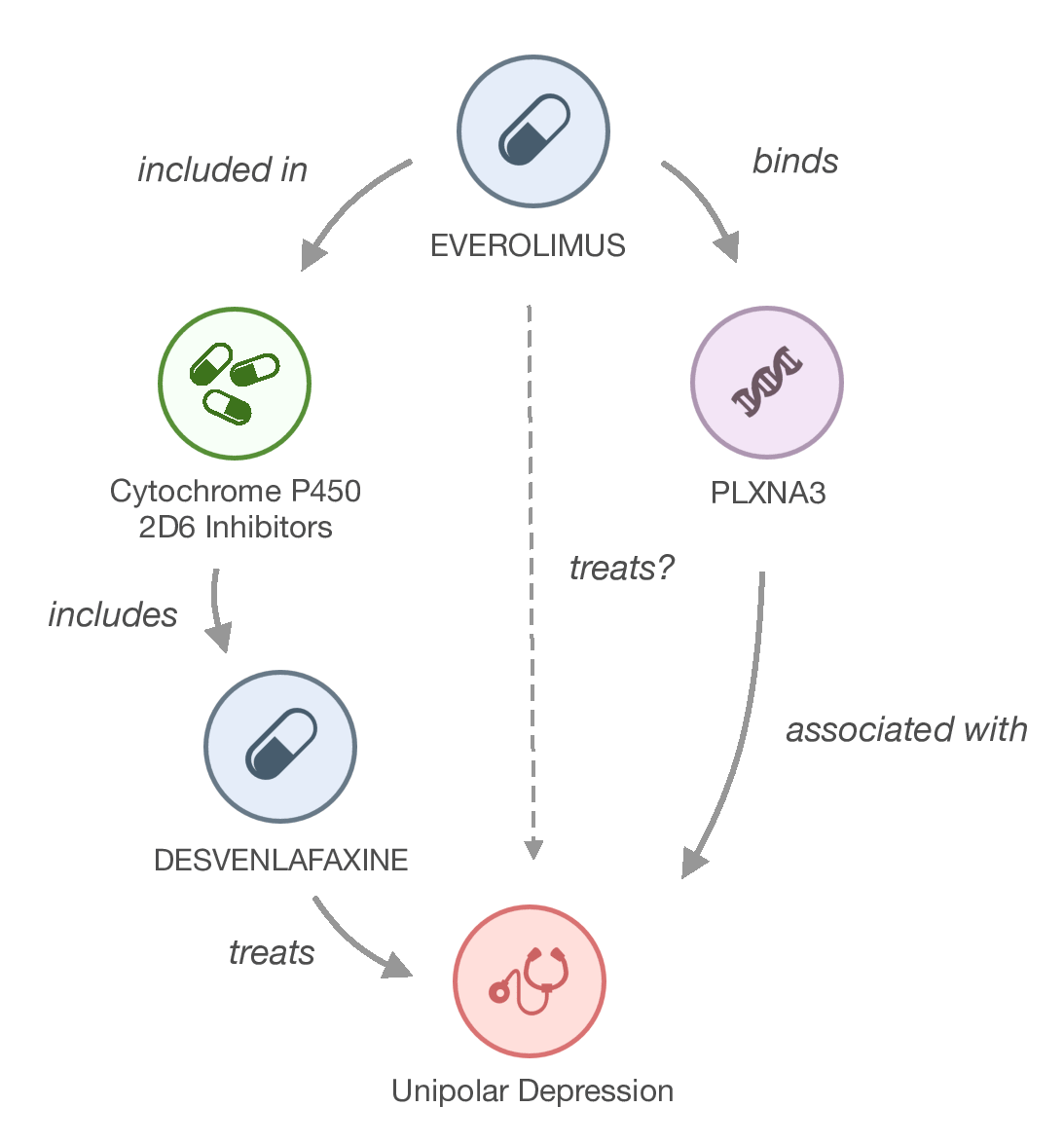}
    \caption{Transparent reasoning for why Everolimus could treat Unipolar Depression.}
    \label{fig:reasoning_path}
\end{figure}

AI may pave the way for innovative new therapies, but pursuing the recommendations means devoting time and resources. First, experts are needed to label and rank recommendation lists, and then laboratory scientists must undertake costly and often time-consuming experiments to provide further validation. Finally, stakeholders need to persevere when provided recommendations are not successful, which is certain to happen, given the notoriously high failure rates of new drugs. The whole process demands robust, sustained trust from humans. Therefore, a human-centric approach is essential. We argue that, in tandem with accurate recommendations, providing transparent reasoning, as shown in Figure \ref{fig:reasoning_path}, is a key part to building the trust required.

Drug discovery recommendation tasks include, but are not limited to, drug repositioning, disease target identification and personalised medicine. Some recent successes have been obtained in the application of recommendation systems on top of networks and knowledge graphs of biological knowledge \citep{barabasi2011network}. For instance, machine learning models trained on graphs have been used to find novel drug candidates for COVID-19 \citep{zhou2020network, gysi2021network}, identify efficacious drug combination therapies \citep{cheng2019network} and discover novel drivers of drug resistance \citep{gogleva2021knowledge}.

A Knowledge Graph (KG) is a data representation that links diverse types of data together into a single, unified model. It enables complex and nuanced relationships to be captured. Due to the increasing breadth and depth of biomedical data, biomedical KGs have become a popular way to integrate the complex and diverse data into a single unified representation.

Biomedical KGs can contain and integrate data ranging from disease networks and protein interactions to healthcare records and scientific knowledge extracted from literature. The intrinsic patterns, uncertainty and bias within biomedical data presents many unique challenges for reasoning \citep{bonner2021review}. Biomedical KGs are also structurally different from standard KGs (Figure \ref{fig:descriptives}). Because of this, the common assumptions and methods for graph reasoning often fail to generalise \citep{liu2021neural}.

\subsection{Main contributions:} This paper makes the following novel contributions:

\begin{enumerate}
    \item Applying multi-hop neural-driven recommendation to complex biomedical KGs 10x bigger than previously used.
    \item Benchmarking multiple algorithms on representative biomedical datasets and tasks with standardised evaluations.
    \item Validating if biologically relevant explanations are produced with biomedical expert end users.
    \item Finding that multi-hop reasoning has the potential to generate explanations and boost the performance of black-box methods.
\end{enumerate}
\section{Related Work}

\subsection{Knowledge Graphs}

\begin{figure}[htbp]
    \centering
\begin{subfigure}{
        \centering
        \begin{tikzpicture}[scale=0.55, every node/.style={scale=0.6}]
            \Vertex[IdAsLabel, Math, size=1.,color=gray, opacity=0.3]{v1}
            \Vertex[IdAsLabel, Math, size=1., x=3,color=gray, opacity=0.3]{v2}
            \Vertex[IdAsLabel, Math, size=1., x=2.5,y=-2,color=gray, opacity=0.3]{v3}
            \Vertex[IdAsLabel, Math, size=1., x=-1.7,y=1.5,color=gray, opacity=0.3]{v4}
            \Vertex[IdAsLabel, Math, size=1., x=-1.9,y=-1.4,color=gray, opacity=0.3]{v5}
            \Vertex[IdAsLabel, Math, size=1., x=1,y=-3.2,color=gray, opacity=0.3]{v6}
            \Vertex[IdAsLabel, Math, size=1., x=.9,y=2.6,color=gray, opacity=0.3]{v7}
            \Edge[lw=2pt](v1)(v3)
            \Edge[lw=2pt](v1)(v4)
            \Edge[lw=2pt](v4)(v7)
            \Edge[lw=2pt](v5)(v4)
            \Edge[lw=2pt](v2)(v7)
            \Edge[lw=2pt](v3)(v6)
            \Edge[lw=2pt](v1)(v6)
            \Edge[lw=2pt](v5)(v6)
        \end{tikzpicture}
        }
    \end{subfigure}
\definecolor{blueish}{RGB}{229,237,249}
\definecolor{pinkish}{RGB}{255,223,219}
\definecolor{greenish}{RGB}{247,255,247}
\definecolor{dark_blueish}{RGB}{104,121,135}
\definecolor{dark_pinkish}{RGB}{217,114,114}
\definecolor{dark_greenish}{RGB}{85,142,57}
\hfill
\begin{subfigure}{
        \centering      
        \begin{tikzpicture}[scale=0.55, every node/.style={scale=0.6}]
            \Vertex[IdAsLabel, Math, size=1., color=pinkish, opacity=1, style={draw=dark_pinkish}, ]{v1_1}
            \Vertex[IdAsLabel, Math, size=1., x=3, color=greenish, opacity=1, style={draw=dark_greenish}, ]{v2_1}
            \Vertex[IdAsLabel, Math, size=1., x=2.5,y=-2,color=blueish, opacity=1, style={draw=dark_blueish},]{v3_1}
            \Vertex[IdAsLabel, Math, size=1., x=-1.7,y=1.5,color=blueish, opacity=1, style={draw=dark_blueish},]{v3_2}
            \Vertex[IdAsLabel, Math, size=1., x=-1.9,y=-1.4, color=greenish, opacity=1, style={draw=dark_greenish},]{v2_2}
            \Vertex[IdAsLabel, Math, size=1., x=1,y=-3.2,color=blueish, opacity=1, style={draw=dark_blueish},]{v3_3}
            \Vertex[IdAsLabel, Math, size=1., x=.9,y=2.6, color=pinkish, opacity=1, style={draw=dark_pinkish},]{v1_2}
            \Edge[Direct, lw=2pt, bend=-10, label=e1, color=red, opacity=0.6](v3_2)(v1_2)
            \Edge[Direct, lw=2pt, bend=-10, label=e2, color=orange, opacity=0.6](v1_2)(v3_2)
            \Edge[Direct, lw=2pt, label=e2, color=orange, opacity=0.6](v1_1)(v3_1)
            \Edge[Direct ,lw=2pt, label=e1, color=red, opacity=0.6](v1_1)(v3_2)
            \Edge[Direct, lw=2pt, label=e1, color=red, opacity=0.6](v2_2)(v3_2)
            \Edge[lw=2pt, label=e3, color=teal, opacity=0.8](v2_1)(v1_2)
            \Edge[lw=2pt,  label=e3, color=teal, opacity=0.8](v3_1)(v3_3)
            \Edge[Direct, lw=2pt, label=e1, color=red, opacity=0.6](v6)(v1_1)
            \Edge[Direct, lw=2pt, label=e2, bend=-10, color=orange, opacity=0.6](v3_3)(v2_2)
            \Edge[Direct, lw=2pt, label=e2, bend=10, color=orange, opacity=0.6](v3_3)(v2_2)
        \end{tikzpicture}
        }
    \end{subfigure}
    \caption{Basic graph (a) vs knowledge graph (b)}
    \label{fig:homo-v-het}
\end{figure}
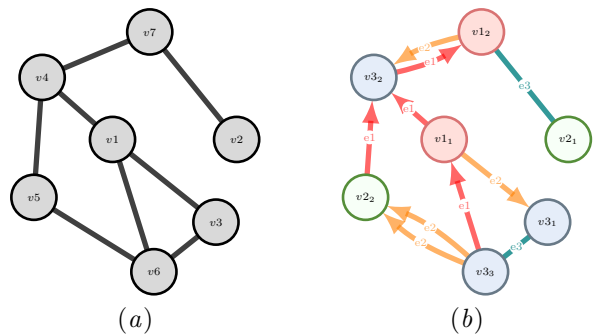

At the fundamental level, graphs contain information about a set of entities (also referred to as nodes or mathematically as vertex) and the set of relationships (also known as edges) between them. A Knowledge Graph (KG) is a heterogeneous, multi-relation graph that can contain directed and undirected edges. This means it has many types of entities and relations, can have multiple edges of the same type between entities, and the edges may indicate direction (Figure \ref{fig:homo-v-het}). 

The core component of a graph is a triple. A single triple defines how two entities are linked. It consists of a head entity $h$, relation $r$ and target entity $t$. A triple is therefore defined by $(h, r, t)$. An example of a triple would be $(London,\text{ } capital\text{ }of,\text{ }United\text{ }Kingdom)$. When multiple triples are combined, it creates a graph. 

A common graph reasoning task is link prediction. It is used to discover new links within incomplete KGs. Many classical AI problems such as recommendations, reasoning and question answering can be rephrased as link prediction problems. The task is to find new links by identifying entities that correctly complete a query. For example, predicting drugs to treat COVID-19 could be defined as finding tail entities that correctly complete the triple $(COVID\text{-}19 , treated\text{ }by, ? )$.

In industry, KGs have been successfully deployed in many applications such as search engines, social graphs, fraud detection and product recommendation. KGs commonly used for benchmarking within research include FB15K-237 \citep{toutanova2015representing}, NELL-995 \citep{xiong2017deeppath} and WN18RR \citep{dettmers2018convolutional}.

However, the lack of large and reliable benchmark graphs for research purposes is well known \citep{hu2020open}. In particular, standard, representative and fair benchmark graphs are still limited for specialist problems such as biomedical recommendation. Initiatives such as the Open Graph Benchmark \citep{hu2020open} are good attempts at solving this, but still not representative of large biomedical KGs used in industry. OGB's biomedical KG is relatively small and has only 5 node types (\appendixref{appendix:dataset_details}). It also differs structurally when compared to the low degree graphs with NLP edges extracted from literature (Figure \ref{fig:descriptives}).

\subsection{Biomedical Knowledge Graphs}
\begin{figure}[htbp]
    \centering
    \includegraphics[scale=0.35]{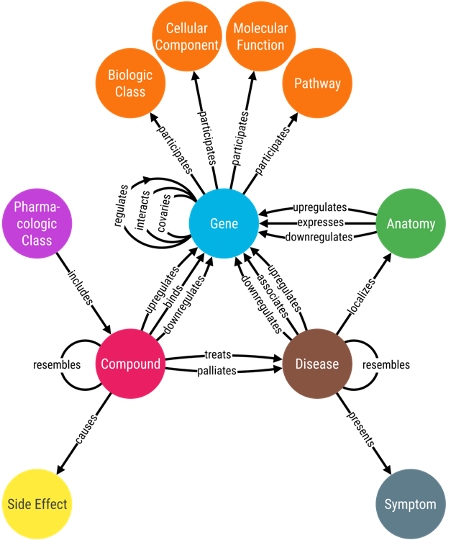}
    \caption{Ontology of Hetionet biomedical knowledge graph \citep{himmelstein2017systematic}. Source: \url{https://het.io/about/} }
    \label{fig:hetionet_schema}
\end{figure}

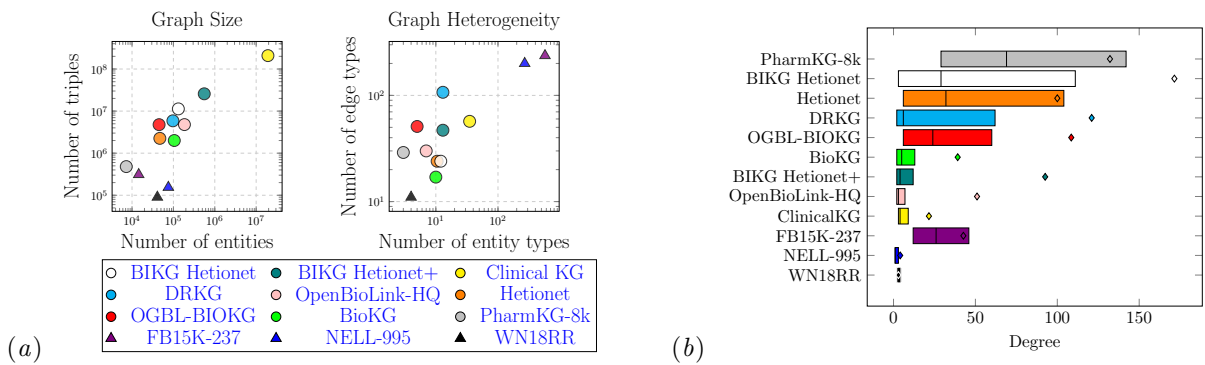
\begin{figure*}[htbp]

\centering
\begin{subfigure}[]{}
\begin{tikzpicture}[scale=0.45,transform shape]
\tikzset{font={\fontsize{16pt}{12}\selectfont}}
	
\begin{groupplot}[group style={
	    group size=2 by 1,
		horizontal sep=90pt,
		vertical sep=70pt,
		ylabels at=edge left},
	width=0.4\textwidth,
	height=0.4\textwidth,
	legend columns=3,
    every tick label/.append style={font=\bf},
    enlarge x limits=true,
	grid=major,
	grid style={dashed, gray!40},
	scaled ticks=false,
	inner axis line style={-stealth}]

\nextgroupplot[
  title=Graph Size,
  xlabel=Number of entities,
  ylabel=Number of triples,
  xmode=log,
  ymode=log,
  ]

\addplot[fill=violet, mark size=5pt, only marks, mark=triangle*, opacity=0.8] coordinates {(14505,310116)};

\addplot[fill=blue, mark size=5pt, only marks, mark=triangle*, opacity=0.8] coordinates {(75492,154213)};

\addplot[fill=black, mark size=5pt, only marks, mark=triangle*, opacity=0.8] coordinates {(40945,90003)};

\addplot[fill=orange, mark size=5pt, only marks, mark=*, opacity=0.8] coordinates {  (47031,2250197)   };

\addplot[fill=pink, mark size=5pt, only marks, mark=*, opacity=0.8] coordinates {  (184667,4778683)   };

\addplot[fill=red, mark size=5pt, only marks, mark=*, opacity=0.8] coordinates {  (45085,4762677)   };

\addplot[fill=lightgray, mark size=5pt, only marks, mark=*, opacity=0.8] coordinates {  (7262,479902)   };

\addplot[fill=cyan, mark size=5pt, only marks, mark=*, opacity=0.8] coordinates {  (97238,5874258)   };

\addplot[fill=yellow, only marks, mark size=5pt, mark=*, opacity=0.8,forget plot] coordinates {(19251579,208177953)};

\addplot[fill=green, only marks, mark size=5pt, mark=*, opacity=0.8,forget plot] coordinates {(105000,2000000)}; 

\addplot[fill=white, only marks, mark size=5pt, mark=*, opacity=0.8,forget plot] coordinates {(131668,11289713)};

\addplot[fill=teal, only marks, mark size=5pt, mark=*, opacity=0.8,forget plot] coordinates {(556219,25762310)};

 \nextgroupplot[
   title=Graph Heterogeneity,
   xlabel=Number of entity types,
   ylabel=Number of edge types,
   legend style = { 
    column sep = 10pt,
    legend columns = 3,
    legend to name = grouplegend
    }, 
    fill opacity=1.0,
    draw opacity=1,
    xmode=log,
    ymode=log,
    ]

\addplot[fill=violet, only marks, mark size=5pt, mark=triangle*, opacity=0.8,forget plot] coordinates {(571,237)};

\addplot[fill=blue, only marks, mark size=5pt, mark=triangle*, opacity=0.8,forget plot] coordinates {(269,200)}; 

\addplot[fill=black, only marks, mark size=5pt, mark=triangle*, opacity=0.8,forget plot] coordinates {(4,11)};

\addplot[fill=orange, only marks, mark size=5pt, mark=*, opacity=0.8,forget plot] coordinates {(10.5,24)}; 

\addplot[fill=pink, only marks, mark size=5pt, mark=*, opacity=0.8,forget plot] coordinates {(7,30)}; 

\addplot[fill=red, only marks, mark size=5pt, mark=*, opacity=0.8,forget plot] coordinates {(5,51)}; %

\addplot[fill=lightgray, only marks, mark size=5pt, mark=*, opacity=0.8,forget plot] coordinates {(3,29)};

\addplot[fill=cyan, only marks, mark size=5pt, mark=*, opacity=0.8,forget plot] coordinates {(13,107)};

\addplot[fill=yellow, only marks, mark size=5pt, mark=*, opacity=0.8,forget plot] coordinates {(35,57)};

\addplot[fill=green, only marks, mark size=5pt, mark=*, opacity=0.8,forget plot] coordinates {(10,17)};

\addplot[fill=white, only marks, mark size=5pt, mark=*, opacity=0.8,forget plot] coordinates {(12,24)};

\addplot[fill=teal, only marks, mark size=5pt, mark=*, opacity=0.8,forget plot] coordinates {(13,47)};

\addlegendimage{mark=*,mark size=4pt,only marks,mark options={fill=white},color=black}
\addlegendentry{BIKG Hetionet}

\addlegendimage{mark=*,mark size=4pt,only marks,mark options={fill=teal},color=black}
\addlegendentry{BIKG Hetionet+}

\addlegendimage{mark=*,mark size=4pt,only marks,mark options={fill=yellow},color=black }
\addlegendentry{Clinical KG}

\addlegendimage{mark=*,mark size=4pt,only marks,mark options={fill=cyan},color=black}
\addlegendentry{DRKG}

\addlegendimage{mark=*,mark size=4pt,only marks,mark options={fill=pink},color=black}
\addlegendentry{OpenBioLink-HQ}

\addlegendimage{mark=*,mark size=4pt,only marks,mark options={fill=orange},color=black}
\addlegendentry{Hetionet}

\addlegendimage{mark=*,mark size=4pt,only marks,mark options={fill=red},color=black}
\addlegendentry{OGBL-BIOKG}

\addlegendimage{mark=*,mark size=4pt,only marks,mark options={fill=green},color=black}
\addlegendentry{BioKG}

\addlegendimage{mark=*,mark size=4pt,only marks,mark options={fill=lightgray},color=black}
\addlegendentry{PharmKG-8k}

\addlegendimage{mark=*,mark size=5pt,only marks,mark options={fill=violet,mark=triangle*},color=black}
\addlegendentry{FB15K-237}

\addlegendimage{mark=*,mark size=5pt,only marks,mark options={fill=blue,mark=triangle*}}
\addlegendentry{NELL-995}

\addlegendimage{mark=*,mark size=5pt,only marks,mark options={fill=black,mark=triangle*}}
\addlegendentry{WN18RR}

\end{groupplot}
    \node at ($(group c1r1) + (4.5cm,-5.3cm)$) {\ref{grouplegend}}; 

\end{tikzpicture}
\end{subfigure}
\qquad %
\begin{subfigure}[]{}
\begin{tikzpicture}[scale=0.65,transform shape]
    \centering
      \begin{axis}
        [
        ytick={1,2,3,4,5,6,7,8,9,10,11,12},
        yticklabels={WN18RR,NELL-995,FB15K-237,ClinicalKG,OpenBioLink-HQ, BIKG Hetionet+,BioKG,OGBL-BIOKG,DRKG,Hetionet,BIKG Hetionet,PharmKG-8k},
        xlabel=Degree,
        area legend,
        ]
        
        \addplot+[
        boxplot prepared={
          median=4.525,
          average=3,
          upper quartile=5,
          lower quartile=2,
        },
        fill=black,
        draw=white,
        solid,
        ] coordinates {};
        
        \addplot+[
        boxplot prepared={
          median=1,
          average=4.085,
          upper quartile=3,
          lower quartile=1,
        },
        fill=blue,
        draw=black,
        solid,
        ] coordinates {};

        \addplot+[
        boxplot prepared={
          median=26,
          average=42.65,
          upper quartile=46,
          lower quartile=12,
        },
        fill=violet,
        draw=black,
        solid,
        ] coordinates {};

        \addplot+[
        boxplot prepared={
          median=4,
          average=21.62693,
          upper quartile=9,
          lower quartile=3,
        },
        fill=yellow,
        draw=black,
        solid,
        ] coordinates {};
        
        \addplot+[
        boxplot prepared={
          median=3,
          average=51,
          upper quartile=7,
          lower quartile=2,
        },
        fill=pink,
        draw=black,
        solid,
        ] coordinates {};

        \addplot+[
        boxplot prepared={
          median=4,
          average=92.6,
          upper quartile=12,
          lower quartile=2,
        },
        fill=teal,
        draw=black,
        solid,
        ] coordinates {};
        
        \addplot+[
        boxplot prepared={
          median=5,
          average=39.19,
          upper quartile=13,
          lower quartile=2,
        },
        fill=green,
        draw=black,
        solid,
        ] coordinates {};

        \addplot+[
        boxplot prepared={
          median=24,
          average=108.52,
          upper quartile=60,
          lower quartile=6,
        },
        fill=red,
        draw=black,
        solid,
        ] coordinates {};
        
        \addplot+[
        boxplot prepared={
          median=6,
          average=121,
          upper quartile=62,
          lower quartile=2,
        },
        fill=cyan,
        draw=black,
        solid,
        ] coordinates {};

        \addplot+[
        boxplot prepared={
          median=32,
          average=100,
          upper quartile=104,
          lower quartile=6,
        },
        fill=orange,
        draw=black,
        solid,
        ] coordinates {};

        \addplot+[
        boxplot prepared={
          median=29,
          average=171.49,
          upper quartile=111,
          lower quartile=3,
        },
        fill=white,
        draw=black,
        solid,
        ] coordinates {};
        
        \addplot+[
        boxplot prepared={
          median=69,
          average=132.16,
          upper quartile=142,
          lower quartile=29,
        },
        fill=lightgray,
        draw=black,
        solid,
        ] coordinates {};

      \end{axis}
    \end{tikzpicture}
\end{subfigure}

	\caption{Comparison of biological and generic KG datasets. \textbf{(a)} \textit{Graph size and heterogeneity.} Triangles ($\Delta$) represent benchmark KGs, circles ($\circ$) note biomedical KGs. \textbf{(b)} \textit{Skewness of degree distribution.}  Box plots represent the quartiles of the degree distributions and the mean degree is denoted by the diamonds ($\diamondsuit$). Biomedical KGs have a highly skewed degree distribution caused by a long-tail of poorly connected entities and a large number of highly connected hub-like entities.}\label{fig:descriptives}
	
\end{figure*}

Biomedical KGs are a specialist type of KG that focus on the biomedical domain. They include information between entities such as genes, diseases, pathways and compounds. The data is usually collected from a wide variety of data sources \citep{bonner2021review}. An example ontology can be seen in Figure \ref{fig:hetionet_schema}. The biomedical data is large, noisy, incomplete and contains contradictory observations. Structurally, the graphs exhibit long-range dependencies, many high-degree hub entities, a higher density of links, long tails of weakly connected entities and higher heterogeneity than standard benchmark graphs (Figure \ref{fig:descriptives}).

Biomedical KGs are becoming increasingly popular within industry and as open-source initiatives. Examples include, BioKG \citep{walsh2020biokg}, DRKG \citep{ioannidis2020drkg}, Clinical KG \citep{santos2020clinical}, Hetionet \citep{himmelstein2017systematic}, OpenBioLink \citep{breit2020openbiolink}, OGBL-BIOKG \citep{hu2020open} and PharmKG \citep{zheng2020pharmkg}.

\subsection{Existing Reasoning Approaches}
Existing approaches to reasoning on KGs originate from different theoretical perspectives with varying levels of explainability. The approaches can be categorised into symbolic, neural and neurosymbolic reasoning approaches \citep{zhang2021neural}.

\paragraph*{Symbolic}
Symbolism assumes symbols are the fundamental unit of human intelligence. It treats cognition purely as a series of inferences upon symbolic representations \citep{haugeland1989artificial}. Symbolic reasoning approaches deduce general logic rules to use when predicting new links.

Importantly, as they follow clear symbolic rules, the reasoning is fully transparent to the user. AnyBURL \citep{meilicke2019anytime} and pLogicNet \citep{qu2019probabilistic} both involve the learning of rules. Recent updates to AnyBURL added a more neuro-symbolic style. They were shown to yield impressive but not superior performance on the biomedical KG Hetionet \citep{liu2021neural}.

Due to the strict matching and discrete logic operations, symbolic methods are thought to be intolerant to the ambiguity and noise of biomedical data. However, the inflexible rules can restrict the expressiveness. Additionally, it was hypothesised that large amounts of high-degree entities makes the learning and applying of logical rules more difficult \citep{liu2021neural}. 

\paragraph*{Neural}
In contrast, connectionism is subsymbolic and imitates the neurons within brains to build models \citep{rosenblatt1958perceptron,rumelhart1986learning}. Instead of symbolic representations, neural reasoning approaches map entities and edges in KGs to low-dimensional vector representations known as embeddings. This approach is the focus of a large body of recent research \citep{wu2020comprehensive}. 

Despite achieving state-of-the-art results on benchmark tasks, a fundamental problem with the neural approach is its non-transparent nature. Predictions are single-hop within vector space and the contributing features driving the prediction are hidden to humans.

Traditional embeddings methods are trained to minimise the reconstruction error in the immediate first-order neighbourhood while discarding higher-order proximities. There is also a homophily assumption in proximity preserving embeddings. However, most explanatory metapaths for drug repurposing were found to have length two or more \citep{liu2021neural} and biomedical graphs display heterophily.

More recently, Graph Neural Networks (GNNs) learn node embeddings by aggregating incoming messages from neighbouring nodes. In theory, these methods are capable of modeling longer-term dependencies. However, in the presence of many high-degree nodes, like in biomedical KGs, the size of neighbourhoods leads to many uninformative signals and causes over-smoothing \citep{kipf2016semi}.

GNNs are also limited by poor expressiveness of long-tail entities with small neighbourhoods \citep{guo2019learning}. This is common in biomedical KGs, demonstrated by the average degree's 25th percentile of just 3 neighbours and positive skews in Figure \ref{fig:descriptives}.

\paragraph*{Neurosymbolic}
The neural-symbolic approach attempts to synergise the neural and symbolic approaches. It is being increasingly referred to by the colloquial term "neurosymbolic". In graph reasoning, this leverages a symbolic graph representation to reduce the search space, cleanse uninformative signals and provide interpretability, whilst using neural approaches to increase expressiveness and adapt the reasoning to ambiguity. 

These attributes are highly relevant for reasoning on biomedical KGs. Multi-hop reasoning with neural networks, a path based approach to KG reasoning, falls into the subcategory of neural-driven symbolic reasoning. 

\subsection{Reinforcement Learning Reasoning}
Formulating KG reasoning as a neural-driven multi-hop problem naturally lends itself to Reinforcement Learning (RL). RL is known to be effective in using neural approaches for solving sequential decision-making problems. Alternative neurosymbolic approaches such as NTP \citep{rocktaschel2017end} and Neural LP \citep{yang2017differentiable} fail to scale to large knowledge graphs \citep{liu2021neural}. 

RL methods have achieved superhuman levels of performance in sequential decision making, famously for arcade games \citep{mnih2013playing}, board games \citep{silver2017mastering}, online multiplayer games \citep{berner2019dota} and imperfect-information games \citep{brown2020combining}. These successes demonstrated RL's ability to operate on ambiguous data, understand complex environments and infer high-level causal relationships. We believe these to be very relevant attributes for the application to biomedical KGs and drug discovery recommendations in general.

The approach of using RL for recommendation systems is increasing in popularity in industry and academia \citep{afsar2021reinforcement, chen2019top}. RL has already been applied to the tasks of question answering \citep{das2018go}, fact-checking \citep{xiong2017deeppath} and recommendations \citep{liu2021neural}.

Most of the approaches use path based RL reasoning, which sequentially traverses triples in a graph to make predictions. Because of this, it provides transparent reasoning for every prediction. An issue with this approach can be convergence and generating relevant reasoning paths. To help, demonstrations of reasoning can be provided by metapaths which acts as a form of weak supervision \citep{zhao2020leveraging, liu2021neural}. RL reasoning applied to the standard benchmarks graphs used in research has not shown promising performance and thus has not been the focus of much research. A small sample of path-based models and their characteristics are shown in Table \ref{table:rl_comparisons}.

\begin{table*}[htbp]
\centering
\small
\caption{Sample of RL reasoning algorithms}
\begin{tabular}{lll}
Algorithm & Reference&Approach \\ \toprule
DeepPath& \cite{xiong2017deeppath}     & RL path finding for fact checking \\
MINERVA& \cite{das2018go}  & Policy-based RL for QA   \\
M-Walk& \cite{shen2018m}   & Monte Carlo Tree Search when sampling             \\
MultiHop& \cite{lin2018multi} & Soft reward and action dropout for MINERVA \\
PGPR&  \cite{xian2019reinforcement} &  Policy-based RL walks       \\
RuleGuider& \cite{lei2020learning}   &   Discovers rules to use for RL walks    \\
R2D2& \cite{hildebrandt2020reasoning}   &   MINERVA with debate dynamics    \\
PoLo &\cite{liu2021neural}   & MINERVA with metapaths
\end{tabular}
\label{table:rl_comparisons}
\end{table*}

MINERVA is a path based RL algorithm for the similar task of question answering. Two extensions of MINERVA, R2D2 and PoLo, applied the method to a biomedical KG \citep{hildebrandt2020reasoning, liu2021neural}. PoLo highlighted the potential for multi-hop reasoning to be especially suited to biomedical KGs.

Whilst promising in their approach, these works used Hetionet which is a relatively small KG that lacks the full complexity, size and noise found in industrial biomedical KGs. Additionally, Hetionet has not been updated since 2017. PoLo's results came from testing a limited number (151) of Compound-treats-Disease links and the evaluation strategy was not standardised as pruning was not tested across all models. The ambiguity of how metrics are calculated also creates the possibility for unfair comparisons \citep{sun2020reevaluation, berrendorf2021ambiguity}. To overcome the limitations of these studies, we extend their work to larger biomedical KGs and offer a set of standardised evaluations.

\section{Preliminaries}

The algorithms MINERVA and PoLo are used for path-based RL reasoning to directly build on the claims in \citet{liu2021neural} that the approach is suited to biomedical KGs. Both algorithms use the same definitions of the knowledge graph environment and task, which are defined below.

\subsection{Knowledge Graph}
A triple links two entities together. It is defined by $(h, r, t)$ where $h$ is the head entity, $r$ is the relation, and $t$ is the tail entity. An example of a triple would be $(Ibuprofen, treats, Headache)$. \set{E} represents the full set of entities and \set{R} the set full of relations in a graph. Every entity has a unique type \set{T}. The KG is defined as a set of $(h,r,t)$ triples $\set{KG} \subset \set{E} \times \set{R} \times \set{E}$. To enable RL agents to travel bidirectionally across triples, the inverse triples are added to the $KG$. The inverse relations are denoted as $r^{-1}$. For example for $(e_{1},r,e_{2})$ the triple $(e_{2},r^{-1},e_{1})$ is also added. All triples in the KG are treated as true facts.

\subsection{Metapaths}
A metapath is a logical reasoning rule defining a general pathway between two entities. For example, a desired metapath for predicting $(Compound, treats, Disease)$ triples might be:
\[
(Compound\overset{binds}{\rightarrow}Gene\overset{associates}{\rightarrow}Disease)
\]

\subsection{Recommendation}
The task of recommendation is posed as a link prediction problem via sequential graph traversal. Given a query $(e_{1}, r, ?)$, such as $(COVID\text{-}19, treated\text{ }by, ? )$, the task is to sequentially traverse triples in the KG to find the correct answer entities. At each step $s$ in the graph, the set of possible actions is the outgoing triples of the current entity $e_{s}$. Each action is represented by the triples relation and tail entity $(r,t)$. Vector space embeddings are used to represent the relations and entities for input to the models. A 'NO\_OP' action is augmented to each node, which effectively allows for no action to be taken and staying at the current node. A maximum number of steps is set to prevent the model from traversing indefinitely.

\section{Experiments}

We ran multiple experiments to establish if the path based RL reasoning approach is suited to biomedical recommendation tasks. We applied it to three different KGs on two separate tasks to draw concrete conclusions on its suitability. Ablations are performed to understand the approach in more depth. 

\subsection{Tasks}

The models are evaluated on two common biomedical recommendation tasks: \emph{drug repurposing} and \emph{drug-target interaction}. In the first task, the goal is to find $(Compound, treats, Disease)$ triples. Formally the query is defined as $(Compound, treats, ? )$. For the second task, the goal is predicting $(Compound, binds, Gene )$ triples, with the formal query being defined as $(Compound, binds, ? )$.

\subsection{Datasets}

Three datasets are used. Hetionet as a small open-source baseline graph and two projections of Biological Insights Knowledge Graph (BIKG) \citep{bikg2021}. BIKG is an internal biomedical KG actively used by AstraZeneca for drug discovery recommendations. The two projections are `BIKG Hetionet' and `BIKG Hetionet+'. The specific projections of BIKG are chosen as they represent the two separate structural types of biomedical KGs, which can be seen in Figure \ref{fig:descriptives}. Importantly, these specific datasets were chosen as they share the same core ontology as Hetionet (Figure \ref{fig:hetionet_schema}), which allows for a unique systematic analysis.

\paragraph*{Hetionet}
Hetionet \citep{himmelstein2017systematic} is an open-source dataset developed in 2017 as part of a study looking at drug repurposing using a KG. It is the most commonly used biomedical graph for benchmarking. Hetionet combines data from 29 public databases and contains over 47k entities of 11 types, including genes, compounds and diseases. These are linked by over 2.2M edges of 24 different relation types. The full ontology is shown in Figure \ref{fig:hetionet_schema}.

\paragraph*{BIKG Hetionet}
BIKG \citep{bikg2021} unifies data from different public, private and internal sources into a single, comprehensive knowledge graph. BIKG Hetionet is a projection of BIKG using the same Hetionet schema in Figure \ref{fig:hetionet_schema}, except with almost five times more edges and three times more nodes. It represents the type of biomedical graph with a higher density and average degree (Figure \ref{fig:descriptives}). Usually this is because they do not include edges mined from literature. Exact dataset statistics can be found in \appendixref{appendix:experiment_details}. 

\paragraph*{BIKG Hetionet+}
Like other biomedical KGs, a large part of the full BIKG is from Natural Language Processing (NLP) pipelines which extract triples from literature. These are inherently more noisy and ambiguous edges. BIKG Hetionet+ extends on BIKG Hetionet by adding `has link' NLP edges and 22 other new edge types. This nearly doubles the number of edge types. One extra node type `Mechanism of Action' is also added. These additions result in a much noisier and more complex KG, representing the sparser biomedical graphs that contain literature edges and a lower average degree caused by a longer tail of poorly connected entities (Figure \ref{fig:descriptives}). Exact statistics can be found in \appendixref{appendix:experiment_details}.

\subsection{Algorithms}

For the RL algorithm, MINERVA \citep{das2018go} is selected. PoLo \citep{liu2021neural}, which extends MINERVA with extra rewards for following pre-defined metapaths, is also used on the drug repurposing task. Using PoLo tests the effect of weak supervision from injecting prior domain knowledge. The metapaths are re-used from the original paper.

The following baseline algorithms were chosen to cover the different theoretical approaches and the most competitive models from the initial \citet{liu2021neural} results on Hetionet -- TransE \citep{wang2014knowledge}, DistMult \citep{yang2014embedding}, R-GCN \citep{schlichtkrull2018modeling} and AnyBURL (JUNO version) \citep{meilicke2019anytime}.

\subsection{Evaluation Strategy}

Ambiguity exists in the evaluation of link prediction methods. Metrics are not uniformly calculated across individual algorithm implementations and common software libraries. This leads to unfair comparisons in reported results \citep{sun2020reevaluation, berrendorf2021ambiguity}. Therefore, we implement a independent and standardised approach to fairly evaluate predictions. To the best of our knowledge, this is the first time it has been done for biomedical KG recommendations across symbolic, neural and neuro-symbolic methods. %

All metrics are filtered \citep{NIPS2013_1cecc7a7}, deduplicated and evaluated for tail-sided predictions against all nodes in the graph. Pruning is also applied which filters predictions to only nodes of the correct target type, as this would be done before giving recommendation lists to scientists. Standard metrics for recommendation tasks are used, HITS@1, HITS@3, HITS@10, and mean reciprocal rank (MRR).

An initial hyperparameter search is applied for every dataset, task and model to find sensible hyperparameters (\appendixref{appendix:experiment_details}). The best hyperparameters are used for training 5-fold cross-validation with seeded randomly split triples. Finally, the mean of the five runs is reported alongside standard errors.

\section{Results}

The results are shown in Table \ref{table:experiment_results}. The interpretable RL methods, MINERVA and PoLo, performed the best. Averaged across all datasets and tasks, MINERVA outperformed the best baselines by 41.8\% for HITS@1, 24.3\% for HITS@3, 7.3\% for HITS@10 and 13.5\% for MRR. Generating a average performance gain of 21.7\% across metrics.

When metapaths were used in PoLo, it improved MINERVA's average performance by a further 4.4\%. We found applying pruning to RL results was especially important; leading to significant performance boosts, as it effectively eliminates incomplete and failed walks from predictions (\appendixref{appendix:pre-pruned_results} for pre-pruned).

The RL approaches superior performance notably decreased as the graphs complexity increased. Falling from 41.7\% performance gain in BIKG Hetionet to just 6.8\% gain in BIKG Hetionet+. The improvement from metapaths also degraded from 5.4\% on BIKG Hetionet to 2.7\% across metrics for BIKG Hetionet+ as the metapaths became less relevant for the expanded ontology.

The baseline neural and symbolic methods varied in performance, whereas neurosymbolic methods (MINERVA and PoLo) were able to perform consistently across datasets and tasks. The embedding models (TransE \& DistMult) struggled with larger biomedical KGs. However, TransE performed notably well for drug-target interaction. R-GCN performed relatively consistently but achieved a middle-of-the-pack performance. Finally, AnyBURL was competitive on larger graphs with the higher ambiguity. The symbolic rules managed to cut through the noise and perform well. However, in other graphs AnyBURL's rigid rules limited its expressiveness.

In summary, we found that neural methods were better suited to smaller, cleaner biomedical KGs and symbolic methods were suited to larger, noisy biomedical KGs. However the neurosymbolic methods were able to perform consistently well across both.  

\subsection{Ablations}

\paragraph*{Impact of Multi-Hop Reasoning}
An ablation was ran to isolate the impact of single-hop vector space reasoning vs multi-hop symbolic reasoning. It was found that using multi-hop symbolic reasoning had better performance (Table \ref{table:ablations}) This ablation boosted the performance of black-box TransE recommendations whilst also providing the much needed explanations. It highlights the potential for multi-hop reasoning to work synergistically with graph representations usually used for single-hop vector space predictions. Multi-hop reasoning algorithms such as MINERVA could be a way to add explainability on top of black-box approaches like GNNs and traditional embeddings without sacrificing performance. However, further research is required into the extent that this is true. 

\paragraph*{Impact of Neural-Driven Reasoning}
A second ablation was run to see how the expressiveness of neural driven reasoning impacts performance. With superior symbolic rules as guidance, the extra expressiveness of neural-driven reasoning could not outperform the purely symbolic rules (Table \ref{table:ablations}). This shows how symbolic rules are very effective when there are high levels of noise and complexity in the KG. Even when being guided by the symbolic rules, the dynamic capability of the neural driven approach did not provide any benefits.

\clearpage
\begin{table*}[th!]
\vspace{-8mm}
\centering
\small
\definecolor{safegreen}{RGB}{0,158,115}
\definecolor{safeblue}{RGB}{86,180,23}
\caption{Experiment results. \colorbox{green!25}{\textbf{green}} indicates the best result, \colorbox{blue!8}{\textit{blue}} indicates second best result. The interpretable, path based RL methods PoLo and MINERVA performed best.}
{\footnotesize
\begin{tabular}{ccccccc}

\textbf{Task}               & \textbf{Dataset}   & \textbf{Model}   & \textbf{HITS@1} & \textbf{HITS@3} & \textbf{HITS@10} & \textbf{MRR} \\ \toprule

\multirow{22}{*}{\textbf{\begin{tabular}[c]{@{}c@{}c@{}}Drug\\Repurposing\end{tabular}}}

& \multirow{6}{*}{\textbf{Hetionet}}
                               & TransE  & $.193\pm.027$      & $.364\pm.031$      & $.608\pm.034$       & $.312\pm.022$   \\[2pt]
                            &   & DistMult  & $.028\pm.002$      & $.138\pm.022$      & $.373\pm.039$       & $.129\pm.002$   \\[2pt]
                            &   & RGCN  & $.047\pm.028$      & $.194\pm.018$      & $.492\pm.043$       & $.221\pm.040$   \\[2pt]
                            &   & AnyBURL  & $.116\pm.013$      & $.288\pm.024$      & $.572\pm.036$       & $.258\pm.012$   \\[2pt]
                            &   & MINERVA & \cellcolor{blue!8}$\mathit{.351\pm.040}$      & \cellcolor{blue!8}$\mathit{.559\pm.058}$      & \cellcolor{blue!8}$\mathit{.790\pm.057}$       & \cellcolor{blue!8}$\mathit{.463\pm.041}$   \\[2pt]
                            &   & PoLo    & \cellcolor{green!25}$\mathbf{.378\pm.023}$      & \cellcolor{green!25}$\mathbf{.575\pm.025}$      & \cellcolor{green!25}$\mathbf{.845\pm.009}$       & \cellcolor{green!25}$\mathbf{.479\pm.015}$   \\ [2pt]\cmidrule{2-7}
& \multirow{6}{*}{\textbf{\begin{tabular}[c]{@{}c@{}}BIKG\\ Hetionet\end{tabular}}} 
                               & TransE  & $.016\pm.003$      & $.034\pm.005$      & $.088\pm.001$       & $.045\pm.004$   \\[2pt]
                            &   & DistMult  & $.004\pm.002$      & $.013\pm.002$      & $.036\pm.007$       & $.024\pm.003$   \\[2pt]
                            &   & RGCN  & $.056\pm.005$      & $.108\pm.008$      & $.239\pm.008$       & $.117\pm.005$   \\[2pt]
                            &   & AnyBURL  & $.000\pm.000$      & $.015\pm.001$      & $.046\pm.004$       & $.023\pm.001$   \\[2pt]
                            &   & MINERVA & \cellcolor{blue!8}$\mathit{.121\pm.002}$      & \cellcolor{blue!8}$\mathit{.194\pm.010}$      & \cellcolor{blue!8}$\mathit{.249\pm.013}$       & \cellcolor{blue!8}$\mathit{.160\pm.004}$   \\[2pt]
                            &   & PoLo    & \cellcolor{green!25}$\mathbf{.124\pm.005}$      & \cellcolor{green!25}$\mathbf{.204\pm.010}$      & \cellcolor{green!25}$\mathbf{.273\pm.016}$       & \cellcolor{green!25}$\mathbf{.167\pm.008}$   \\ [2pt]\cmidrule{2-7}
& \multirow{6}{*}{\textbf{\begin{tabular}[c]{@{}c@{}}BIKG\\ Hetionet+\end{tabular}}} 
                                & TransE  & $.017\pm.002$      & $.052\pm.004$      & $.119\pm.011$       & $.057\pm.002$   \\[2pt]
                            &   & DistMult  & $.004\pm.002$      & $.014\pm.004$      & $.039\pm.001$       & $.026\pm.004$   \\[2pt]
                            &   & RGCN  & $.039\pm.008$      & $.104\pm.016$      & \cellcolor{green!25}$\mathbf{.222\pm.015}$       & \cellcolor{green!25}$\mathbf{.115\pm.005}$   \\[2pt]
                           &    & AnyBURL  & $.000\pm.000$      & $.056\pm.006$      & \cellcolor{blue!8}$\mathit{.205\pm.014}$       & $.072\pm.002$  \\[2pt]
                           &    & MINERVA & \cellcolor{green!25}$\mathbf{.076\pm.008}$      & \cellcolor{blue!8}$\mathit{.124\pm.011}$      & $.166\pm.013$       & \cellcolor{blue!8}$\mathit{.113\pm.010}$   \\[2pt]
                           &    & PoLo    & \cellcolor{blue!8}$\mathit{.074\pm.006}$      & \cellcolor{green!25}$\mathbf{.133\pm.010}$      & $.173\pm.012$       & \cellcolor{green!25}$\mathbf{.115\pm.007}$  \\[2pt]\midrule

\multirow{19}{*}{\textbf{\begin{tabular}[c]{@{}c@{}}Drug-Target\\Interaction\end{tabular}}}
& \multirow{6}{*}{\textbf{Hetionet}}
                               & TransE  & \cellcolor{green!25}$\mathbf{.287\pm.008}$      & \cellcolor{green!25}$\mathbf{.546\pm.013}$      & \cellcolor{green!25}$\mathbf{.918\pm.007}$       & \cellcolor{green!25}$\mathbf{.331\pm.005}$   \\[2pt]
                            &  & DistMult  & $.039\pm.004$      & $.128\pm.007$      & $.367\pm.012$       & $.115\pm.005$   \\[2pt]
                            &  & RGCN  & $.060\pm.025$      & $.177\pm.057$      & $.376\pm.102$       & $.126\pm.031$   \\[2pt]
                            &  & AnyBURL  & $.169\pm.043$      & $.323\pm.081$      & $.493\pm.124$       & $.210\pm.051$   \\[2pt]
                            &  & MINERVA & \cellcolor{blue!8}$.186\pm.015$      & \cellcolor{blue!8}$\mathit{.420\pm.028}$      & \cellcolor{blue!8}$\mathit{.807\pm.039}$       & \cellcolor{blue!8}$\mathit{.296\pm.019}$   \\[2pt]\cmidrule{2-7}
& \multirow{6}{*}{\textbf{\begin{tabular}[c]{@{}c@{}}BIKG\\ Hetionet\end{tabular}}} 
                               & TransE  & $.193\pm.001$      & $.376\pm.012$      & \cellcolor{blue!8}$\mathit{.674\pm.012}$       & $.239\pm.004$   \\[2pt]
                            &  & DistMult  & $.044\pm.007$      & $.111\pm.014$      & $.269\pm.014$       & $.098\pm.007$   \\[2pt]
                            &  & RGCN  & $.173\pm.016$      & $.338\pm.013$      & $.539\pm.021$       & $.235\pm.008$   \\[2pt]
                            &  & AnyBURL  & \cellcolor{blue!8}$\mathit{.215\pm.005}$      & \cellcolor{blue!8}$\mathit{.408\pm.004}$      & $.623\pm.009$       & \cellcolor{blue!8}$\mathit{.264\pm.004}$   \\[2pt]
                            &  & MINERVA & \cellcolor{green!25}$\mathbf{.235\pm.005}$      & \cellcolor{green!25}$\mathbf{.516\pm.011}$      & \cellcolor{green!25}$\mathbf{.983\pm.015}$       & \cellcolor{green!25}$\mathbf{.305\pm.004}$   \\[2pt]\cmidrule{2-7}
& \multirow{6}{*}{\textbf{\begin{tabular}[c]{@{}c@{}}BIKG\\ Hetionet+\end{tabular}}}          
                               & TransE  & $.054\pm.004$      & $.098\pm.007$      & $.171\pm.011$       & $.008\pm.004$   \\[2pt]
                            &  & DistMult  & $.035\pm.006$      & $.084\pm.011$      & $.206\pm.029$       & $.084\pm.001$   \\[2pt]
                            &  & RGCN  & \cellcolor{blue!8}$\mathit{.188\pm.008}$      & $.323\pm.019$      & $.539\pm.010$       & $.241\pm.006$   \\[2pt]
                            &  & AnyBURL  & \cellcolor{green!25}$\mathbf{.215\pm.003}$      & \cellcolor{green!25}$\mathbf{.408\pm.008}$      & \cellcolor{blue!8}$\mathit{.622\pm.005}$       & \cellcolor{green!25}$\mathbf{.263\pm.003}$   \\[2pt]
                            &  & MINERVA & $.181\pm.006$      & \cellcolor{blue!8}$\mathit{.368\pm.008}$      & \cellcolor{green!25}$\mathbf{.628\pm.019}$       & \cellcolor{blue!8}$\mathit{.243\pm.004}$   \\[2pt]
                            
\end{tabular}
}
\label{table:experiment_results}

\end{table*}

\begin{table*}[th!]
\centering
\small
\caption{\textit{Multi-Hop Ablation.} Retraining MINERVA using the same embeddings from TransE where they initially performed best (Drug-Target Interaction on Hetionet) improved performance. \textit{Neural-Driven Ablation.} Where AnyBURLs symbolic rules performed best (Drug-Target Interaction on BIKG Hetionet+) the same rules were reused as metapaths in PoLo but it failed to improve performance.}
\begin{tabular}{cccccc}
Ablation               & Model    & HITS@1  & HITS@3  & HITS@10 & MRR     \\ \toprule
\multirow{3}{*}{Multi-Hop}     & TransE   & \cellcolor{blue!8}$\mathit{.287\pm.008}$ & \cellcolor{blue!8}$\mathit{.546\pm.013}$ & \cellcolor{blue!8}$\mathit{.918\pm.007}$ & \cellcolor{blue!8}$\mathit{.331\pm.005}$ \\
                               & MINERVA  & $.186\pm.015$ & $.420\pm.028$ & $.807\pm.039$ & $.296\pm.019$ \\
                               & MINERVA* & \cellcolor{green!25}$\mathbf{.327\pm.014}$ & \cellcolor{green!25}$\mathbf{.622\pm.013}$ & \cellcolor{green!25}$\mathbf{.938\pm.018}$ & \cellcolor{green!25}$\mathbf{.350\pm.006}$ \\ \midrule
\multirow{3}{*}{Neural-Driven} & AnyBURL   & \cellcolor{green!25} $\mathbf{.215\pm.003}$ & \cellcolor{green!25} $\mathbf{.408\pm.008}$ & \cellcolor{blue!8}$\mathit{.622\pm.005}$ & \cellcolor{green!25} $\mathbf{.263\pm.003}$ \\
                               & MINERVA  & $.181\pm.006$ & \cellcolor{blue!8}$\mathit{.368\pm.008}$ & \cellcolor{green!25} $\mathbf{.628\pm.019}$ & $.243\pm.004$ \\
                               & PoLo*    & \cellcolor{blue!8}$\mathit{.191\pm.009}$ & $.366\pm.012$ & $.596\pm.019$ & \cellcolor{blue!8}$\mathit{.250\pm.010}$                               
\end{tabular}
\label{table:ablations}
\vspace{-2mm}%
\end{table*}
\clearpage

\section{Explanations}
Although the results are promising, \citet{lv2021multi} claim the reasoning produced by multi-hop models can have gaps in interpretability. If the explanations are unclear to the end user, it will be detrimental to trust. To address this aspect, we run inference on the full BIKG Hetionet and BIKG Hetionet+ graphs, using the trained MINERVA models. We then assess the resulting high-level metapaths and individual reasoning paths for biological relevance.

\subsection{Reasoning Metapaths}
\begin{figure}[htbp!] 
    \centering 
    \includegraphics[scale=0.14]{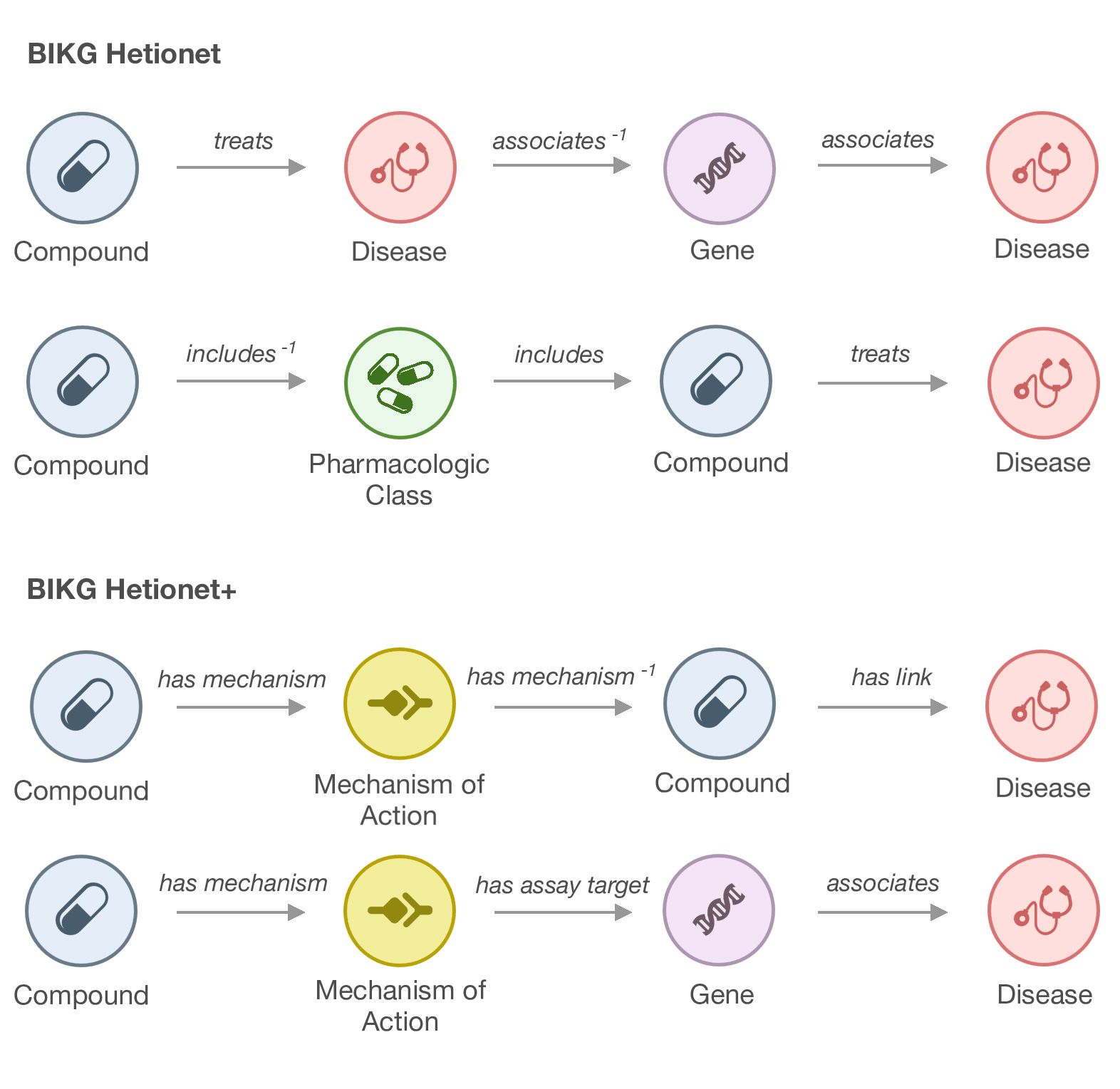}
    \caption{MINERVA's most common metapaths for drug repurposing}
    \label{fig:repurposing_explanations}
\end{figure}

\begin{figure}[htbp!]
    \centering 
    \includegraphics[scale=0.14]{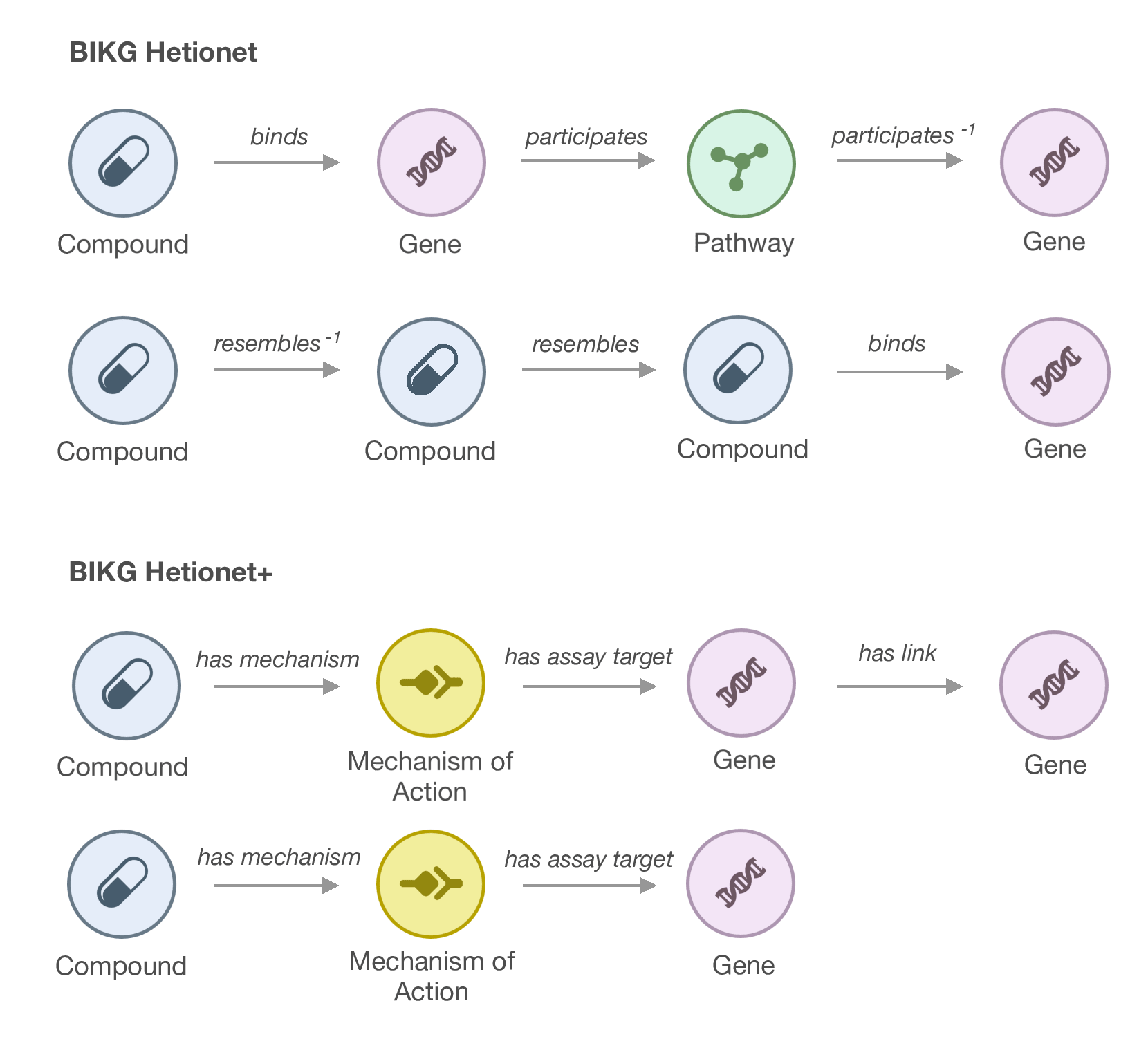}
    \caption{MINERVA's most common metapaths for drug-target interaction}
    \label{fig:drug-target_explanations}
\end{figure}

We found that the high-level metapaths displayed sensible biological logic. The reasoning ranges from clear interactions between genes, mechanisms of action and biological pathways to more vague similarity-based reasons such as compounds that resemble each other (Figure \ref{fig:repurposing_explanations} \& \ref{fig:drug-target_explanations}). However, inspection of the metapaths also revealed that models can get fixated on less successful policies, almost exclusively focusing on mechanisms of action entities in BIKG Hetionet+ and using more cyclic paths, which ultimately led to lower performance. The top 10 metapaths and their frequency can be found in \appendixref{appendix:explanations}. 

\subsection{Case Studies}

Although the high-level metapaths appear to be relevant, the low-level reasoning could still be biologically flawed. Therefore, together with subject matter experts we investigated a sample of the individual reasoning paths to check the validity.

Novel predictions were made for Lixisenatide, Dulaflutide and similar compounds to treat non-alcoholic fatty liver disease (NAFLD). It was able to correctly reason that they act as a receptor agonist (RA) for the gene Glucagon-like peptide-1 receptor (GLP1R). There is literature supporting the role of GLP-1 RA’s as a treatment for NAFLD \citep{seghieri2018future, lv2020glucagon}, as well as clinical trials and case studies for NAFLD treatment \citep{gluud2014effects, seko2017effect}. The next highest prediction for NAFLD was for Secukinumab, which follows a similar reasoning but instead via Interleukin-17A inhibition. There is a clinical trial currently recruiting to study this on ClinicalTrials.gov (NCT04237116).

Everolimus, usually used to treat cancer, was unconventionally predicted to treat unipolar depression (Figure \ref{fig:reasoning_path}). However there is case study evidence to support this theory \citep{mir2018everolimus}. The reasoning was due to links with Cytochrome P450, which has known links to depression via the mTor pathway. It also links to Gene PLXNA3, which is important for neurological development and has studies linking it to depression \citep{wray2007anxiety}. This is a more speculative, but still biologically relevant, reasoning that would require additional validation work from drug discovery scientists.

Finally, a prediction was made of Simvastatin binding to gene UGT1A1. This example highlights how the predictions and explanations are susceptible to errors in the KG data. The reasoning used the edges of Simvastatin binding to enzymatic genes CYP2D6, SLCO1B1 and CYP3A5, which are involved in drug metabolic pathways.UGT1A1 was also shown to be a participant in these pathways, and therefore it reasoned that Simvastatin was likely to bind to UGT1A1. All these enzymatic genes are indeed involved in Simvastatin metabolism \citep{iwuchukwu2014genetic}. However, they do not actually 'bind' to these genes, the edges are misclassified and should be 'affects'. The benefit of the transparent explanations is that a scientist can easily identify this fault and dismiss the recommendation. It also means data issues can easily be identified and fixed, which creates higher quality KG datasets. Whereas for black-box recommendations the fault in the reasoning would be invisible to end users and likely go unnoticed.

These case studies are typical of the types of biological reasoning that can be found throughout the predictions, with some novel and interesting explanations. There is also a proportion of predictions with more ambiguity, linking entities together via similar genes, compounds and pharmacological classes. We provide more examples in \appendixref{appendix:explanations}.

\section{Implications}

The superior performance we observed and the accompanying transparent explanations make the RL approach well suited to biomedical recommendations. The general approach can be extended to produce even greater explanations. RL's reward function enables users to have more control over the reasoning used and it's possible to provide reasons a recommendation might not be true  \citep{hildebrandt2020reasoning}. 

Additionally, the multi-hop approach allows for an interactive, human-in-the-loop reasoning process. Humans and AI can collaborate by combining knowledge to collaboratively walk across the knowledge graph to make predictions. All these attributes mean the approach can create better informed, more trusted biomedical recommendations without compromising on accuracy.

\section{Conclusions}
This paper highlights that biomedical KG recommendation is a unique challenge that cannot be approached in the same way as other recommendation tasks. The reality of the problem demands recommendations to be human-centric and it is shown that biomedical KGs unique structure differentiates them as a distinct problem. Different theoretical types of algorithms were evaluated across multiple representative datasets and tasks. It was found that multi-hop neural-driven symbolic reasoning, using RL, outperformed other approaches. Additionally, using this approach we were able to produce novel and relevant biological explanations. An ablation also highlighted the potential for this approach to create explanations for black-box models whilst boosting performance. Therefore, we believe the approach is well suited to biomedical recommendations and encourage more research in this direction.

\clearpage

\section*{Acknowledgments}
The authors would like to thank Fatima Lugtu, Greet De Baets and Piotr Grabowski for their assistance in evaluating the recommendations and explanations. Thanks also goes to the SCP (Scientific Compute Platform) team at AstraZeneca.

\bibliography{sherlock}

\begin{thebibliography}{65}
\providecommand{\natexlab}[1]{#1}
\providecommand{\url}[1]{\texttt{#1}}
\expandafter\ifx\csname urlstyle\endcsname\relax
  \providecommand{\doi}[1]{doi: #1}\else
  \providecommand{\doi}{doi: \begingroup \urlstyle{rm}\Url}\fi

\bibitem[Abadi et~al.(2016)Abadi, Barham, Chen, Chen, Davis, Dean, Devin,
  Ghemawat, Irving, Isard, et~al.]{abadi2016tensorflow}
Mart{\'\i}n Abadi, Paul Barham, Jianmin Chen, Zhifeng Chen, Andy Davis, Jeffrey
  Dean, Matthieu Devin, Sanjay Ghemawat, Geoffrey Irving, Michael Isard, et~al.
\newblock {TensorFlow: A System for Large-Scale Machine Learning}.
\newblock In \emph{12th $\{$USENIX$\}$ Symposium on Operating Systems Design
  and Implementation ($\{$OSDI$\}$ 16)}, pages 265--283, 2016.

\bibitem[Afsar et~al.(2021)Afsar, Crump, and Far]{afsar2021reinforcement}
M~Mehdi Afsar, Trafford Crump, and Behrouz Far.
\newblock Reinforcement learning based recommender systems: A survey.
\newblock \emph{arXiv preprint arXiv:2101.06286}, 2021.

\bibitem[Akiba et~al.(2019)Akiba, Sano, Yanase, Ohta, and
  Koyama]{akiba2019optuna}
Takuya Akiba, Shotaro Sano, Toshihiko Yanase, Takeru Ohta, and Masanori Koyama.
\newblock Optuna: A next-generation hyperparameter optimization framework.
\newblock In \emph{Proceedings of the 25th ACM SIGKDD International Conference
  on Knowledge Discovery \& Data mining}, pages 2623--2631, 2019.

\bibitem[Austin and Hayford(2021)]{austin_hayford_2021}
David Austin and Tamara Hayford.
\newblock Research and development in the pharmaceutical industry, Apr 2021.
\newblock URL \url{https://www.cbo.gov/publication/57025}.

\bibitem[Barab{\'a}si et~al.(2011)Barab{\'a}si, Gulbahce, and
  Loscalzo]{barabasi2011network}
Albert-L{\'a}szl{\'o} Barab{\'a}si, Natali Gulbahce, and Joseph Loscalzo.
\newblock {Network Medicine: a Network-based Approach to Human Disease}.
\newblock \emph{{Nature Reviews Genetics}}, 12\penalty0 (1):\penalty0 56--68,
  2011.

\bibitem[Berner et~al.(2019)Berner, Brockman, Chan, Cheung, Debiak, Dennison,
  Farhi, Fischer, Hashme, Hesse, et~al.]{berner2019dota}
Christopher Berner, Greg Brockman, Brooke Chan, Vicki Cheung, Przemys{\l}aw
  Debiak, Christy Dennison, David Farhi, Quirin Fischer, Shariq Hashme, Chris
  Hesse, et~al.
\newblock Dota 2 with large scale deep reinforcement learning.
\newblock \emph{arXiv preprint arXiv:1912.06680}, 2019.

\bibitem[Berrendorf et~al.(2021)Berrendorf, Faerman, Vermue, and
  Tresp]{berrendorf2021ambiguity}
Max Berrendorf, Evgeniy Faerman, Laurent Vermue, and Volker Tresp.
\newblock {On the Ambiguity of Rank-Based Evaluation of Entity Alignment or
  Link Prediction Methods}, 2021.

\bibitem[Bonner et~al.(2021)Bonner, Barrett, Ye, Swiers, Engkvist, Bender,
  Hoyt, and Hamilton]{bonner2021review}
Stephen Bonner, Ian~P Barrett, Cheng Ye, Rowan Swiers, Ola Engkvist, Andreas
  Bender, Charles~Tapley Hoyt, and William Hamilton.
\newblock A review of biomedical datasets relating to drug discovery: A
  knowledge graph perspective, 2021.

\bibitem[Bordes et~al.(2013)Bordes, Usunier, Garcia-Duran, Weston, and
  Yakhnenko]{NIPS2013_1cecc7a7}
Antoine Bordes, Nicolas Usunier, Alberto Garcia-Duran, Jason Weston, and Oksana
  Yakhnenko.
\newblock {Translating Embeddings for Modeling Multi-relational Data}.
\newblock In C.~J.~C. Burges, L.~Bottou, M.~Welling, Z.~Ghahramani, and K.~Q.
  Weinberger, editors, \emph{Advances in Neural Information Processing
  Systems}, volume~26. Curran Associates, Inc., 2013.

\bibitem[Breit et~al.(2020)Breit, Ott, Agibetov, and
  Samwald]{breit2020openbiolink}
Anna Breit, Simon Ott, Asan Agibetov, and Matthias Samwald.
\newblock {OpenBioLink: A Benchmarking Framework for Large-Scale Biomedical
  Link Prediction}.
\newblock \emph{Bioinformatics}, 36\penalty0 (13):\penalty0 4097--4098, 2020.

\bibitem[Brown et~al.(2020)Brown, Bakhtin, Lerer, and Gong]{brown2020combining}
Noam Brown, Anton Bakhtin, Adam Lerer, and Qucheng Gong.
\newblock Combining deep reinforcement learning and search for
  imperfect-information games.
\newblock \emph{arXiv preprint arXiv:2007.13544}, 2020.

\bibitem[Chen et~al.(2019)Chen, Beutel, Covington, Jain, Belletti, and
  Chi]{chen2019top}
Minmin Chen, Alex Beutel, Paul Covington, Sagar Jain, Francois Belletti, and
  Ed~H Chi.
\newblock Top-k off-policy correction for a reinforce recommender system.
\newblock In \emph{Proceedings of the Twelfth ACM International Conference on
  Web Search and Data Mining}, pages 456--464, 2019.

\bibitem[Cheng et~al.(2019)Cheng, Kov{\'a}cs, and
  Barab{\'a}si]{cheng2019network}
Feixiong Cheng, Istv{\'a}n~A Kov{\'a}cs, and Albert-L{\'a}szl{\'o}
  Barab{\'a}si.
\newblock {Network-based Prediction of Drug Combinations}.
\newblock \emph{{Nature Communications}}, 10\penalty0 (1):\penalty0 1--11,
  2019.

\bibitem[Das et~al.(2018)Das, Dhuliawala, Zaheer, Vilnis, Durugkar,
  Krishnamurthy, Smola, and McCallum]{das2018go}
Rajarshi Das, Shehzaad Dhuliawala, Manzil Zaheer, Luke Vilnis, Ishan Durugkar,
  Akshay Krishnamurthy, Alex Smola, and Andrew McCallum.
\newblock {Go for a Walk and Arrive at the Answer: Reasoning Over Paths in
  Knowledge Bases Using Reinforcement Learning}.
\newblock In \emph{International Conference on Learning Representations}, 2018.
\newblock URL \url{https://openreview.net/forum?id=Syg-YfWCW}.

\bibitem[Dettmers et~al.(2018)Dettmers, Minervini, Stenetorp, and
  Riedel]{dettmers2018convolutional}
Tim Dettmers, Pasquale Minervini, Pontus Stenetorp, and Sebastian Riedel.
\newblock {Convolutional 2D Knowledge Graph Embeddings}.
\newblock In \emph{{Thirty-Second AAAI Conference on Artificial Intelligence}},
  2018.

\bibitem[Fey and Lenssen(2019)]{fey2019fast}
Matthias Fey and Jan~Eric Lenssen.
\newblock {Fast graph representation learning with PyTorch Geometric}.
\newblock \emph{arXiv preprint arXiv:1903.02428}, 2019.

\bibitem[Geleta et~al.(2021)Geleta, Nikolov, Edwards, Gogleva, Jackson,
  Jansson, Lamov, Nilsson, Pettersson, Poroshin, Rozemberczki, Scrivener,
  Ughetto, and Papa]{bikg2021}
David Geleta, Andriy Nikolov, Gavin Edwards, Anna Gogleva, Richard Jackson,
  Erik Jansson, Andrej Lamov, Sebastian Nilsson, Marina Pettersson, Vladimir
  Poroshin, Benedek Rozemberczki, Timothy Scrivener, Michael Ughetto, and
  Eliseo Papa.
\newblock Biological insights knowledge graph: an integrated knowledge graph to
  support drug development, 2021.
\newblock URL \url{https://doi.org/10.1101/2021.10.28.466262}.

\bibitem[Gluud et~al.(2014)Gluud, Knop, and Vilsb{\o}ll]{gluud2014effects}
Lise~L Gluud, Filip~K Knop, and Tina Vilsb{\o}ll.
\newblock Effects of lixisenatide on elevated liver transaminases: systematic
  review with individual patient data meta-analysis of randomised controlled
  trials on patients with type 2 diabetes.
\newblock \emph{BMJ open}, 4\penalty0 (12):\penalty0 e005325, 2014.

\bibitem[Gogleva et~al.(2021)Gogleva, Polychronopoulos, Pfeifer, Poroshin,
  Ughetto, Sidders, Dry, Ahdesm{\"a}ki, McDermott, Papa,
  et~al.]{gogleva2021knowledge}
Anna Gogleva, Dimitris Polychronopoulos, Matthias Pfeifer, Vladimir Poroshin,
  Micha{\"e}l Ughetto, Ben Sidders, Jonathan Dry, Miika Ahdesm{\"a}ki, Ultan
  McDermott, Eliseo Papa, et~al.
\newblock Knowledge graph-based recommendation framework identifies novel
  drivers of resistance in egfr mutant non-small cell lung cancer.
\newblock \emph{bioRxiv}, 2021.

\bibitem[Guo et~al.(2019)Guo, Sun, and Hu]{guo2019learning}
Lingbing Guo, Zequn Sun, and Wei Hu.
\newblock Learning to exploit long-term relational dependencies in knowledge
  graphs.
\newblock In \emph{International Conference on Machine Learning}, pages
  2505--2514. PMLR, 2019.

\bibitem[Gysi et~al.(2021)Gysi, Do~Valle, Zitnik, Ameli, Gan, Varol, Ghiassian,
  Patten, Davey, Loscalzo, et~al.]{gysi2021network}
Deisy~Morselli Gysi, {\'I}talo Do~Valle, Marinka Zitnik, Asher Ameli, Xiao Gan,
  Onur Varol, Susan~Dina Ghiassian, JJ~Patten, Robert~A Davey, Joseph Loscalzo,
  et~al.
\newblock {Network Medicine Framework for Identifying Drug-Repurposing
  Opportunities for COVID-19}.
\newblock \emph{Proceedings of the National Academy of Sciences}, 118\penalty0
  (19), 2021.

\bibitem[Haugeland(1989)]{haugeland1989artificial}
John Haugeland.
\newblock \emph{Artificial intelligence: The very idea}.
\newblock MIT press, 1989.

\bibitem[Hildebrandt et~al.(2020)Hildebrandt, Serna, Ma, Ringsquandl, Joblin,
  and Tresp]{hildebrandt2020reasoning}
Marcel Hildebrandt, Jorge Andres~Quintero Serna, Yunpu Ma, Martin Ringsquandl,
  Mitchell Joblin, and Volker Tresp.
\newblock Reasoning on knowledge graphs with debate dynamics.
\newblock In \emph{Proceedings of the AAAI Conference on Artificial
  Intelligence}, volume~34, pages 4123--4131, 2020.

\bibitem[Himmelstein et~al.(2017)Himmelstein, Lizee, Hessler, Brueggeman, Chen,
  Hadley, Green, Khankhanian, and Baranzini]{himmelstein2017systematic}
Daniel~Scott Himmelstein, Antoine Lizee, Christine Hessler, Leo Brueggeman,
  Sabrina~L Chen, Dexter Hadley, Ari Green, Pouya Khankhanian, and Sergio~E
  Baranzini.
\newblock {Systematic Integration of Biomedical Knowledge Prioritizes Drugs for
  Repurposing}.
\newblock \emph{Elife}, 6:\penalty0 e26726, 2017.

\bibitem[Hu et~al.(2020)Hu, Fey, Zitnik, Dong, Ren, Liu, Catasta, and
  Leskovec]{hu2020open}
Weihua Hu, Matthias Fey, Marinka Zitnik, Yuxiao Dong, Hongyu Ren, Bowen Liu,
  Michele Catasta, and Jure Leskovec.
\newblock {Open Graph Benchmark: Datasets for Machine Learning on Graphs}.
\newblock \emph{arXiv preprint arXiv:2005.00687}, 2020.

\bibitem[Ioannidis et~al.(2020)Ioannidis, Song, Manchanda, Li, Pan, Zheng,
  Ning, Zeng, and Karypis]{ioannidis2020drkg}
Vassilis~N Ioannidis, Xiang Song, Saurav Manchanda, Mufei Li, Xiaoqin Pan,
  Da~Zheng, Xia Ning, Xiangxiang Zeng, and George Karypis.
\newblock {DRKG - Drug Repurposing Knowledge Graph for COVID-19}, 2020.

\bibitem[Iwuchukwu et~al.(2014)Iwuchukwu, Feng, Wei, Jiang, Jiang, Xu, Denny,
  Wilke, Krauss, Roden, et~al.]{iwuchukwu2014genetic}
Otito~F Iwuchukwu, QiPing Feng, Wei-Qi Wei, Lan Jiang, Min Jiang, Hua Xu,
  Joshua~C Denny, Russell~A Wilke, Ronald~M Krauss, Dan~M Roden, et~al.
\newblock Genetic variation in the ugt1a locus is associated with simvastatin
  efficacy in a clinical practice setting.
\newblock \emph{Pharmacogenomics}, 15\penalty0 (14):\penalty0 1739--1747, 2014.

\bibitem[Kingma and Ba(2017)]{kingma2017adam}
Diederik~P. Kingma and Jimmy Ba.
\newblock Adam: A method for stochastic optimization, 2017.

\bibitem[Kipf and Welling(2016)]{kipf2016semi}
Thomas~N Kipf and Max Welling.
\newblock Semi-supervised classification with graph convolutional networks.
\newblock \emph{arXiv preprint arXiv:1609.02907}, 2016.

\bibitem[Lei et~al.(2020)Lei, Jiang, Gu, Sun, Mao, and Ren]{lei2020learning}
Deren Lei, Gangrong Jiang, Xiaotao Gu, Kexuan Sun, Yuning Mao, and Xiang Ren.
\newblock {Learning Collaborative Agents with Rule Guidance for Knowledge Graph
  Reasoning}.
\newblock In \emph{Proceedings of the 2020 Conference on Empirical Methods in
  Natural Language Processing (EMNLP)}, pages 8541--8547, 2020.

\bibitem[Lin et~al.(2018)Lin, Socher, and Xiong]{lin2018multi}
Xi~Victoria Lin, Richard Socher, and Caiming Xiong.
\newblock Multi-hop knowledge graph reasoning with reward shaping.
\newblock \emph{arXiv preprint arXiv:1808.10568}, 2018.

\bibitem[Liu et~al.(2021)Liu, Hildebrandt, Joblin, Ringsquandl, Raissouni, and
  Tresp]{liu2021neural}
Yushan Liu, Marcel Hildebrandt, Mitchell Joblin, Martin Ringsquandl, Rime
  Raissouni, and Volker Tresp.
\newblock {Neural Multi-Hop Reasoning With Logical Rules on Biomedical
  Knowledge Graphs}.
\newblock In \emph{European Semantic Web Conference}, pages 375--391. Springer,
  2021.

\bibitem[Lv et~al.(2020)Lv, Dong, Hu, Lu, Zhou, and Qin]{lv2020glucagon}
Xiaodan Lv, Yongqiang Dong, Lingling Hu, Feiyu Lu, Changyu Zhou, and Shaoyou
  Qin.
\newblock Glucagon-like peptide-1 receptor agonists (glp-1 ras) for the
  management of nonalcoholic fatty liver disease (nafld): A systematic review.
\newblock \emph{Endocrinology, diabetes \& metabolism}, 3\penalty0
  (3):\penalty0 e00163, 2020.

\bibitem[Lv et~al.(2021)Lv, Cao, Hou, Li, Liu, Zhang, and Dai]{lv2021multi}
Xin Lv, Yixin Cao, Lei Hou, Juanzi Li, Zhiyuan Liu, Yichi Zhang, and Zelin Dai.
\newblock Is multi-hop reasoning really explainable? towards benchmarking
  reasoning interpretability.
\newblock \emph{arXiv preprint arXiv:2104.06751}, 2021.

\bibitem[Meilicke et~al.(2019)Meilicke, Chekol, Ruffinelli, and
  Stuckenschmidt]{meilicke2019anytime}
Christian Meilicke, Melisachew~Wudage Chekol, Daniel Ruffinelli, and Heiner
  Stuckenschmidt.
\newblock {Anytime Bottom-Up Rule Learning for Knowledge Graph Completion.}
\newblock In \emph{IJCAI}, pages 3137--3143, 2019.

\bibitem[Mir et~al.(2018)Mir, Salvador, Dauchy, Ropert, Lemogne, and
  Gaillard]{mir2018everolimus}
Olivier Mir, Alexandre Salvador, Sarah Dauchy, Stanislas Ropert, C{\'e}dric
  Lemogne, and Rapha{\"e}l Gaillard.
\newblock Everolimus induced mood changes in breast cancer patients: a
  case-control study.
\newblock \emph{Investigational new drugs}, 36\penalty0 (3):\penalty0 503--508,
  2018.

\bibitem[Mnih et~al.(2013)Mnih, Kavukcuoglu, Silver, Graves, Antonoglou,
  Wierstra, and Riedmiller]{mnih2013playing}
Volodymyr Mnih, Koray Kavukcuoglu, David Silver, Alex Graves, Ioannis
  Antonoglou, Daan Wierstra, and Martin Riedmiller.
\newblock Playing atari with deep reinforcement learning.
\newblock \emph{arXiv preprint arXiv:1312.5602}, 2013.

\bibitem[Nair and Hinton(2010)]{nair2010rectified}
Vinod Nair and Geoffrey~E Hinton.
\newblock Rectified linear units improve restricted boltzmann machines.
\newblock In \emph{Icml}, 2010.

\bibitem[Paszke et~al.(2017)Paszke, Gross, Chintala, Chanan, Yang, DeVito, Lin,
  Desmaison, Antiga, and Lerer]{paszke2017automatic}
Adam Paszke, Sam Gross, Soumith Chintala, Gregory Chanan, Edward Yang, Zachary
  DeVito, Zeming Lin, Alban Desmaison, Luca Antiga, and Adam Lerer.
\newblock Automatic differentiation in pytorch.
\newblock \emph{NIPS}, 2017.

\bibitem[Qu and Tang(2019)]{qu2019probabilistic}
Meng Qu and Jian Tang.
\newblock Probabilistic logic neural networks for reasoning.
\newblock \emph{arXiv preprint arXiv:1906.08495}, 2019.

\bibitem[Rockt{\"a}schel and Riedel(2017)]{rocktaschel2017end}
Tim Rockt{\"a}schel and Sebastian Riedel.
\newblock End-to-end differentiable proving.
\newblock \emph{arXiv preprint arXiv:1705.11040}, 2017.

\bibitem[Rosenblatt(1958)]{rosenblatt1958perceptron}
Frank Rosenblatt.
\newblock The perceptron: a probabilistic model for information storage and
  organization in the brain.
\newblock \emph{Psychological review}, 65\penalty0 (6):\penalty0 386, 1958.

\bibitem[Rumelhart et~al.(1986)Rumelhart, Hinton, and
  Williams]{rumelhart1986learning}
David~E Rumelhart, Geoffrey~E Hinton, and Ronald~J Williams.
\newblock Learning representations by back-propagating errors.
\newblock \emph{nature}, 323\penalty0 (6088):\penalty0 533--536, 1986.

\bibitem[Santos et~al.(2020)Santos, Cola{\c{c}}o, Nielsen, Niu, Geyer, Coscia,
  Albrechtsen, Mundt, Jensen, and Mann]{santos2020clinical}
Alberto Santos, Ana~Rita Cola{\c{c}}o, Annelaura~Bach Nielsen, Lili Niu,
  Philipp~Emanuel Geyer, Fabian Coscia, Nicolai Jacob~Wewer Albrechtsen, Filip
  Mundt, Lars~Juhl Jensen, and Matthias Mann.
\newblock {Clinical Knowledge Graph Integrates Proteomics Data Into Clinical
  Decision-Making}.
\newblock \emph{bioRxiv}, 2020.

\bibitem[Schlichtkrull et~al.(2018)Schlichtkrull, Kipf, Bloem, Van Den~Berg,
  Titov, and Welling]{schlichtkrull2018modeling}
Michael Schlichtkrull, Thomas~N Kipf, Peter Bloem, Rianne Van Den~Berg, Ivan
  Titov, and Max Welling.
\newblock {Modeling Relational Data with Graph Convolutional Networks}.
\newblock In \emph{{European Semantic Web Conference}}, pages 593--607.
  Springer, 2018.

\bibitem[Seghieri et~al.(2018)Seghieri, Christensen, Andersen, Solini, Knop,
  and Vilsb{\o}ll]{seghieri2018future}
Marta Seghieri, Alexander~S Christensen, Andreas Andersen, Anna Solini, Filip~K
  Knop, and Tina Vilsb{\o}ll.
\newblock Future perspectives on glp-1 receptor agonists and glp-1/glucagon
  receptor co-agonists in the treatment of nafld.
\newblock \emph{Frontiers in endocrinology}, 9:\penalty0 649, 2018.

\bibitem[Seko et~al.(2017)Seko, Sumida, Tanaka, Mori, Taketani, Ishiba, Hara,
  Okajima, Umemura, Nishikawa, et~al.]{seko2017effect}
Yuya Seko, Yoshio Sumida, Saiyu Tanaka, Kojiroh Mori, Hiroyoshi Taketani,
  Hiroshi Ishiba, Tasuku Hara, Akira Okajima, Atsushi Umemura, Taichiro
  Nishikawa, et~al.
\newblock Effect of 12-week dulaglutide therapy in japanese patients with
  biopsy-proven non-alcoholic fatty liver disease and type 2 diabetes mellitus.
\newblock \emph{Hepatology Research}, 47\penalty0 (11):\penalty0 1206--1211,
  2017.

\bibitem[Shen et~al.(2018)Shen, Chen, Huang, Guo, and Gao]{shen2018m}
Yelong Shen, Jianshu Chen, Po-Sen Huang, Yuqing Guo, and Jianfeng Gao.
\newblock {M-Walk: Learning to Walk Over Graphs Using Monte Carlo Tree Search}.
\newblock In \emph{Proceedings of the 32nd International Conference on Neural
  Information Processing Systems}, pages 6787--6798, 2018.

\bibitem[Silver et~al.(2017)Silver, Schrittwieser, Simonyan, Antonoglou, Huang,
  Guez, Hubert, Baker, Lai, Bolton, et~al.]{silver2017mastering}
David Silver, Julian Schrittwieser, Karen Simonyan, Ioannis Antonoglou, Aja
  Huang, Arthur Guez, Thomas Hubert, Lucas Baker, Matthew Lai, Adrian Bolton,
  et~al.
\newblock Mastering the game of go without human knowledge.
\newblock \emph{nature}, 550\penalty0 (7676):\penalty0 354--359, 2017.

\bibitem[Srivastava et~al.(2014)Srivastava, Hinton, Krizhevsky, Sutskever, and
  Salakhutdinov]{srivastava2014dropout}
Nitish Srivastava, Geoffrey Hinton, Alex Krizhevsky, Ilya Sutskever, and Ruslan
  Salakhutdinov.
\newblock Dropout: a simple way to prevent neural networks from overfitting.
\newblock \emph{The journal of machine learning research}, 15\penalty0
  (1):\penalty0 1929--1958, 2014.

\bibitem[Sun et~al.(2020)Sun, Vashishth, Sanyal, Talukdar, and
  Yang]{sun2020reevaluation}
Zhiqing Sun, Shikhar Vashishth, Soumya Sanyal, Partha Talukdar, and Yiming
  Yang.
\newblock {A Re-evaluation of Knowledge Graph Completion Methods}, 2020.

\bibitem[Toutanova et~al.(2015)Toutanova, Chen, Pantel, Poon, Choudhury, and
  Gamon]{toutanova2015representing}
Kristina Toutanova, Danqi Chen, Patrick Pantel, Hoifung Poon, Pallavi
  Choudhury, and Michael Gamon.
\newblock {Representing Text for Joint Embedding of Text and Knowledge Bases}.
\newblock In \emph{{Proceedings of the 2015 Conference on Empirical Methods in
  Natural Language Processing}}, pages 1499--1509, 2015.

\bibitem[Walsh et~al.(2020)Walsh, Mohamed, and
  Nov{\'a}{\v{c}}ek]{walsh2020biokg}
Brian Walsh, Sameh~K Mohamed, and V{\'\i}t Nov{\'a}{\v{c}}ek.
\newblock {BioKG: A Knowledge Graph for Relational Learning On Biological
  Data}.
\newblock In \emph{Proceedings of the 29th ACM International Conference on
  Information \& Knowledge Management}, pages 3173--3180, 2020.

\bibitem[Wang et~al.(2014)Wang, Zhang, Feng, and Chen]{wang2014knowledge}
Zhen Wang, Jianwen Zhang, Jianlin Feng, and Zheng Chen.
\newblock {Knowledge Graph Embedding by Translating on Hyperplanes}.
\newblock In \emph{Proceedings of the AAAI Conference on Artificial
  Intelligence}, volume~28, 2014.

\bibitem[Wray et~al.(2007)Wray, James, Mah, Nelson, Andrews, Sullivan,
  Montgomery, Birley, Braun, and Martin]{wray2007anxiety}
Naomi~R Wray, Michael~R James, Steven~P Mah, Matthew Nelson, Gavin Andrews,
  Patrick~F Sullivan, Grant~W Montgomery, Andrew~J Birley, Andreas Braun, and
  Nicholas~G Martin.
\newblock Anxiety and comorbid measures associated with plxna2.
\newblock \emph{Archives of general psychiatry}, 64\penalty0 (3):\penalty0
  318--326, 2007.

\bibitem[Wu et~al.(2020)Wu, Pan, Chen, Long, Zhang, and
  Philip]{wu2020comprehensive}
Zonghan Wu, Shirui Pan, Fengwen Chen, Guodong Long, Chengqi Zhang, and S~Yu
  Philip.
\newblock A comprehensive survey on graph neural networks.
\newblock \emph{IEEE transactions on neural networks and learning systems},
  32\penalty0 (1):\penalty0 4--24, 2020.

\bibitem[Xian et~al.(2019)Xian, Fu, Muthukrishnan, De~Melo, and
  Zhang]{xian2019reinforcement}
Yikun Xian, Zuohui Fu, Shan Muthukrishnan, Gerard De~Melo, and Yongfeng Zhang.
\newblock {Reinforcement Knowledge Graph Reasoning for Explainable
  Recommendation}.
\newblock In \emph{{Proceedings of the 42nd International ACM SIGIR Conference
  on Research and Development in Information Retrieval}}, pages 285--294, 2019.

\bibitem[Xiong et~al.(2017)Xiong, Hoang, and Wang]{xiong2017deeppath}
Wenhan Xiong, Thien Hoang, and William~Yang Wang.
\newblock {DeepPath: A Reinforcement Learning Method for Knowledge Graph
  Reasoning}.
\newblock In \emph{Proceedings of the 2017 Conference on Empirical Methods in
  Natural Language Processing}, pages 564--573, 2017.

\bibitem[Yang et~al.(2014)Yang, Yih, He, Gao, and Deng]{yang2014embedding}
Bishan Yang, Wen-tau Yih, Xiaodong He, Jianfeng Gao, and Li~Deng.
\newblock Embedding entities and relations for learning and inference in
  knowledge bases.
\newblock \emph{arXiv preprint arXiv:1412.6575}, 2014.

\bibitem[Yang et~al.(2017)Yang, Yang, and Cohen]{yang2017differentiable}
Fan Yang, Zhilin Yang, and William~W Cohen.
\newblock Differentiable learning of logical rules for knowledge base
  reasoning.
\newblock \emph{arXiv preprint arXiv:1702.08367}, 2017.

\bibitem[Zhang et~al.(2021)Zhang, Chen, Zhang, Ke, and Ding]{zhang2021neural}
Jing Zhang, Bo~Chen, Lingxi Zhang, Xirui Ke, and Haipeng Ding.
\newblock {Neural, Symbolic and Neural-symbolic Reasoning on Knowledge Graphs}.
\newblock \emph{AI Open}, 2:\penalty0 14--35, 2021.

\bibitem[Zhao et~al.(2020)Zhao, Wang, Zhang, Zhao, Liu, Xing, and
  Xie]{zhao2020leveraging}
Kangzhi Zhao, Xiting Wang, Yuren Zhang, Li~Zhao, Zheng Liu, Chunxiao Xing, and
  Xing Xie.
\newblock {Leveraging Demonstrations for Reinforcement Recommendation Reasoning
  Over Knowledge Graphs}.
\newblock In \emph{Proceedings of the 43rd International ACM SIGIR Conference
  on Research and Development in Information Retrieval}, pages 239--248, 2020.

\bibitem[Zheng et~al.(2020{\natexlab{a}})Zheng, Song, Ma, Tan, Ye, Dong, Xiong,
  Zhang, and Karypis]{zheng2020dgl}
Da~Zheng, Xiang Song, Chao Ma, Zeyuan Tan, Zihao Ye, Jin Dong, Hao Xiong, Zheng
  Zhang, and George Karypis.
\newblock {DGL-KE: Training Knowledge Graph Embeddings at Scale}.
\newblock In \emph{Proceedings of the 43rd International ACM SIGIR Conference
  on Research and Development in Information Retrieval}, pages 739--748,
  2020{\natexlab{a}}.

\bibitem[Zheng et~al.(2020{\natexlab{b}})Zheng, Rao, Song, Zhang, Xiao, Fang,
  Yang, and Niu]{zheng2020pharmkg}
Shuangjia Zheng, Jiahua Rao, Ying Song, Jixian Zhang, Xianglu Xiao, Evandro~Fei
  Fang, Yuedong Yang, and Zhangming Niu.
\newblock {PharmKG: A Dedicated Knowledge Graph Benchmark for Biomedical Data
  Mining}.
\newblock \emph{Briefings in Bioinformatics}, 2020{\natexlab{b}}.

\bibitem[Zhou et~al.(2020)Zhou, Hou, Shen, Huang, Martin, and
  Cheng]{zhou2020network}
Yadi Zhou, Yuan Hou, Jiayu Shen, Yin Huang, William Martin, and Feixiong Cheng.
\newblock {Network-based Drug Repurposing for Novel Coronavirus
  2019-nCoV/SARS-CoV-2}.
\newblock \emph{Cell Discovery}, 6\penalty0 (1):\penalty0 1--18, 2020.

\end{thebibliography}

\clearpage
\appendix

\onecolumn
\section{Experimental Details}
\label{appendix:experiment_details}

\subsection{Model Implementations}

Our experiments with PoLo and MINERVA utilized the TensorFlow \citep{abadi2016tensorflow} codebase released by the authors of PoLo. AnyBURL \citep{meilicke2019anytime} uses the latest version (AnyBURL-JUNO) from the authors. The DistMult and TransE (L2 norm) embeddings were learned with the DGL-KE \citep{zheng2020dgl} on top of the PyTorch \citep{paszke2017automatic} automatic differentiation backend library. The hyperparameters of these embedding models were tuned on the validation set with the open-source Optuna framework \citep{akiba2019optuna}. Finally, the R-GCN baselines were generated with the PyTorch Geometric library \citep{fey2019fast} using Adam optimiser \cite{kingma2017adam}, Dropout \cite{srivastava2014dropout} and ReLU \cite{nair2010rectified}.

\subsection{Hyperparameters}

To get the best results for each model, the Table X and Y show hyperparameters that are searched. For MINERVA, $\lambda$ is set to 0 which recovers MINERVA from PoLo.

\begin{table}[h!]
\centering

\caption{Hyperparameter search space for MINERVA and PoLo. For MINERVA $\lambda$ is set to 0.}

\begin{tabular}{ccc}
Hyperparameter        & Values            \\ \toprule
Embedding size        & \{128, 256\} \\
Hidden layer size     & \{256,512\}   \\
Learning rate         & \{0.0001,0.001\}     \\
$\lambda$             & \{0.1,1\}   \\
$\beta$               & \{0.01,0.1\}    \\
\end{tabular}
\end{table}

\begin{table}[h!]
\centering

\caption{Hyperparameter search space for knowledge graph embedding techniques TransE and DistMult.}

\begin{tabular}{ccc}
Hyperparameter        & Values            \\ \toprule
Embedding size        & \{128, 256, \dots, 1024\} \\
Learning rate         & {[}0.001,0.2{]}     \\
Negative samples      & \{1,\dots,500\}   \\
Max step      & \{12000, 22000, \dots, 200000\}   \\
\end{tabular}
\end{table}

\begin{table}[h!]
\centering

\caption{Hyperparameter search space for R-GCN.}

\begin{tabular}{ccc}
Hyperparameter        & Values            \\ \toprule
Embedding size        & $\{32,64,128\}$ \\
Learning rate         & $\{0.01,0.05\}$     \\
Negative samples      & $\{10\}$   \\
Max step      & $\{0.01\}$   \\

\end{tabular}
\end{table}

\paragraph{AnyBURL}

We use the most recent (JUNO) version of AnyBURL, which is more of a neurosymbolic approach as it implements Reinforcement Learning to learn rules. The maximum length of rules is set to 3 and the rules are learned for a total of 500 seconds. The AnyBURL algorithm does not need hyperparameter tuning to achieve good results, therefore we kept the default values of remaining hyperparameters.

\clearpage
\subsection{Dataset Statistics}
\label{appendix:dataset_details}

\begin{table}[h!]
\centering

\caption{Statistics of each tasks dataset sizes}
\begin{tabular}{cccccccc}
Task                              & Dataset        & Relation & Nodes  & Edges    & Train   & Valid & Test  \\ \toprule
\multirow{3}{*}{Drug Repurposing} & Hetionet       & CtD                          & 47,031  & 2,250,197  & 483     & 121   & 151   \\
                                  & BIKG Hetionet  & CtD                          & 131,668 & 11,289,713 & 2,062    & 516   & 664   \\
                                  & BIKG Hetionet+ & CtD                          & 556,219 & 25,762,310 & 2,114    & 529   & 661   \\ \midrule
\multirow{3}{*}{Drug-Target Interaction}      & Hetionet       & CbG                          & 47,031  & 2,250,197  & 7,404    & 1,852  & 2,315  \\
                                  & BIKG Hetionet  & CbG                         & 131,668 & 11,289,713 & 7,351    & 1,838  & 2,298  \\
                                  & BIKG Hetionet+ & CbG                         & 556,219 & 25,762,310 & 7,351    & 1,838  & 2,298  \\
\end{tabular}
\end{table}

\begin{table}[h!]
\centering
\caption{Statistics of biomedical and benchmark knowledge graphs}

\begin{tabular}{ccccccc}
Dataset        & Domain   & Node Types  & Nodes      & Edge Types & Edges     & Mean Degree \\ \toprule
ClinicalKG     & Biological  & 35       & 19,251,579  & 57         & 208,177,953 & 21.63       \\
BIKG           & Biological & 27        & 10,948,027 & 52        & 83,849,252  & 17.87     \\ %
BIKG Hetionet+ & Biological & 12        & 556,219    & 47         & 25,762,310  & 92.63       \\
BIKG Hetionet  & Biological & 11        & 131,668    & 24         & 11,289,713   & 171.49     \\
DRKG           & Biological & 13        & 97,238     & 107        & 5,874,258   & 120.81      \\
OGBL-BIOKG     & Biological & 5         & 45,085     & 51         & 4,762,677   & 108.52      \\
OpenBioLink-HQ & Biological & 7         & 184,667    & 30         & 4,778,683   & 51.80       \\
Hetionet       & Biological & 11        & 47,031     & 24         & 2,250,197   & 95.83       \\
BioKG          & Biological & 10        & 105,524    & 17         & 2,067,998   & 39.19       \\
PharmKG-8k     & Biological & 3         & 7,262      & 29         & 479,902    & 132.16      \\
FB15K-237      & General    & 571       & 14,505     & 237        & 310,116    & 42.50       \\
WN18RR         & Lexical    & 4         & 40,945     & 11         & 90,003     & 4.53        \\
NELL-995       & General    & 269       & 75,492     & 200        & 154,213    & 4.09        
\end{tabular}
\label{table:dataset_stats}

\end{table}

\clearpage
\section{Explanations}
\label{appendix:explanations}

The following section contains metapath statistics and explanations generated by MINVERA on the BIKG Hetionet and BIKG Hetionet+ datasets.

\subsection{Metapaths Statistics}
\begin{table}[!htbp]
\vspace{-8mm}
\centering
\small
\caption{MINERVA's top 10 frequent reasoning metapaths for drug repurposing (pruned)}

\begin{tabular}{ccc}
\toprule
    Dataset & Metapath &  \% \\
\midrule
\multirow{12}{*}{\textbf{\begin{tabular}[c]{@{}c@{}}BIKG\\ Hetionet\end{tabular}}}
&  $Compound \xrightarrow{\text{treats}} Disease \xrightarrow{\text{associates}^{-1}} Gene \xrightarrow{\text{associates}} Disease$ &       71.7 \\
&  $Compound \xrightarrow{\text{includes}^{-1}} Pharmacologic\ Class \xrightarrow{\text{includes}} Compound \xrightarrow{\text{treats}} Disease$ &       5.1 \\
&   $Compound \xrightarrow{\text{treats}} Disease \xrightarrow{\text{upregulates}} Gene \xrightarrow{\text{associates}} Disease$ &       4.3 \\
&    $Compound \xrightarrow{\text{treats}} Disease \xrightarrow{\text{treats}^{-1}} Compound \xrightarrow{\text{treats}} Disease$ &       2.7 \\
&   $Compound \xrightarrow{\text{palliates}} Disease \xrightarrow{\text{upregulates}} Gene \xrightarrow{\text{associates}} Disease$ &       2.1 \\
&    $Compound \xrightarrow{\text{resembles}} Compound \xrightarrow{\text{resembles}} Compound  \xrightarrow{\text{treats}} Disease$ &       1.5 \\
& $ Compound \xrightarrow{\text{binds}} Gene \xrightarrow{\text{associates}} Disease$ &       1.2 \\
&    $Compound \xrightarrow{\text{resembles}} Compound \xrightarrow{\text{resembles}^{-1}} Compound \xrightarrow{\text{treats}} Disease$ &       1.2 \\
&  $Compound \xrightarrow{\text{palliates}} Disease \xrightarrow{\text{associates}^{-1}} Gene \xrightarrow{\text{associates}} Disease$ &       1.2 \\
&    $Compound \xrightarrow{\text{treats}} Disease \xrightarrow{\text{localizes}} Anatomy \xrightarrow{\text{localizes}^{-1}} Disease'$ &       0.9 \\
\midrule
\multirow{12}{*}{\textbf{\begin{tabular}[c]{@{}c@{}}BIKG\\ Hetionet+\end{tabular}}} 
&   $Compound \xrightarrow{\text{has mechanism}} Mechanism\ of\ Action \xrightarrow{\text{has mechanism}^{-1}} Compound \xrightarrow{\text{has link}} Disease$ &       51.5 \\
&  $Compound \xrightarrow{\text{has mechanism}} Mechanism\ of\ Action \xrightarrow{\text{has assay target}} Gene \xrightarrow{\text{associates}} Disease$ &       34.2 \\
&  $Compound \xrightarrow{\text{has link}} Disease \xrightarrow{\text{has link}^{-1}} Disease \xrightarrow{\text{has link}} Disease$ &       9.2 \\
& $Compound \xrightarrow{\text{has mechanism}} Mechanism\ of\ Action \xrightarrow{\text{has assay target}} Gene \xrightarrow{\text{has link}} Disease$ &       1.6 \\
&        $Compound \xrightarrow{\text{treats}} Disease \xrightarrow{\text{has link}^{-1}} Compound  \xrightarrow{\text{has link}} Disease$ &       1.3 \\
&        $Compound \xrightarrow{\text{treats}} Disease \xrightarrow{\text{associates}^{-1}} Gene \xrightarrow{\text{associates}} Disease$ &       0.4 \\
&         $Compound \xrightarrow{\text{resembles}} Compound  \xrightarrow{\text{binds}} Gene \xrightarrow{\text{associates}} Disease$ &       0.3 \\
&     $Compound \xrightarrow{\text{has link}} Disease \xrightarrow{\text{involved in}^{-1}} Biological\ Process \xrightarrow{\text{involved in}} Disease$ &       0.2 \\
&        $Compound \xrightarrow{\text{resembles}^{-1}} Compound  \xrightarrow{\text{binds}} Gene \xrightarrow{\text{associates}} Disease$ &       0.2 \\
&      $Compound \xrightarrow{\text{treats}} Disease \xrightarrow{\text{involved in}^{-1}} Biological\ Process \xrightarrow{\text{involved in}} Disease$ &       0.1 \\
\bottomrule
\end{tabular}

\end{table}
\begin{table}[!htbp]
\vspace{-8mm}
\centering
\small
\caption{MINERVA's top 10 frequent reasoning metapaths for drug target interaction (pruned)}

\begin{tabular}{ccc}
\toprule
    Dataset & Metapath &  \% \\
\midrule
\multirow{12}{*}{\textbf{\begin{tabular}[c]{@{}c@{}}BIKG\\ Hetionet\end{tabular}}}
&   $Compound \xrightarrow{\text{binds}} Gene \xrightarrow{\text{participates}} Pathway \xrightarrow{\text{participates}^{-1}} Gene$ &       14.2 \\
&     $Compound \xrightarrow{\text{resembles}^{-1}} Compound \xrightarrow{\text{resembles}} Compound \xrightarrow{\text{binds}} Gene$ &       13.8 \\
&  $Compound \xrightarrow{\text{includes}^{-1}} Pharmacologic\ Class \xrightarrow{\text{includes}} Compound \xrightarrow{\text{binds}} Gene$ &       13.2 \\
&      $Compound \xrightarrow{\text{resembles}}  Compound \xrightarrow{\text{resembles}} Compound \xrightarrow{\text{binds}} Gene$ &       12.4 \\
&     $Compound \xrightarrow{\text{resembles}}  Compound \xrightarrow{\text{resembles}^{-1}} Compound \xrightarrow{\text{binds}} Gene$ &       12.2 \\
& $Compound \xrightarrow{\text{binds}}  Gene \xrightarrow{\text{participates}} Molecular\ Function \xrightarrow{\text{participates}^{-1}} Gene$ &       4.8 \\
&    $Compound \xrightarrow{\text{resembles}} Compound \xrightarrow{\text{binds}} Gene$ &       3.5 \\
&    $Compound \xrightarrow{\text{resembles}} Compound \xrightarrow{\text{binds}} Gene$ &       3.3 \\
& $Compound \xrightarrow{\text{binds}} Gene  \xrightarrow{\text{participates}} Biological\ Process \xrightarrow{\text{participates}} Gene$ &       3.2 \\
&    $Compound \xrightarrow{\text{resembles}^{-1}} Compound \xrightarrow{\text{resembles}^{-1}} Compound \xrightarrow{\text{binds}} Gene$ &       3.2 \\
\midrule
\multirow{12}{*}{\textbf{\begin{tabular}[c]{@{}c@{}}BIKG\\ Hetionet+\end{tabular}}} 
& $Compound \xrightarrow{\text{has mechanism}} Mechanism\ of\ Action \xrightarrow{\text{has assay target}} Gene  \xrightarrow{\text{has link}} Gene$ &       47.6 \\
&  $Compound \xrightarrow{\text{has mechanism}} Mechanism\ of\ Action \xrightarrow{\text{has assay target}} Gene$ &       18.8 \\
&  $Compound \xrightarrow{\text{has mechanism}} Mechanism\ of\ Action \xrightarrow{\text{has assay target}} Gene \xrightarrow{\text{interacts}} Gene$ &       10.1 \\
&     $Compound \xrightarrow{\text{binds}} Gene \xrightarrow{\text{has link}} Gene \xrightarrow{\text{has link}} Gene$ &       5.9 \\
&    $Compound \xrightarrow{\text{has link}} Gene \xrightarrow{\text{has link}} Gene \xrightarrow{\text{has link}} Gene$ &       3.0 \\
&      $Compound \xrightarrow{\text{binds}} Gene \xrightarrow{\text{has link}} Gene$ &       2.2 \\
&         $Compound \xrightarrow{\text{associates}} Pathway \xrightarrow{\text{involved in}} Disease \xrightarrow{\text{associates}^{-1}} Gene$ &       1.4 \\
&       $Compound \xrightarrow{\text{resembles}^{-1}} Compound \xrightarrow{\text{binds}} Gene \xrightarrow{\text{has link}} Gene$ &       1.4 \\
&     $Compound \xrightarrow{\text{upregulates}} Gene \xrightarrow{\text{has link}} Gene \xrightarrow{\text{has link}} Gene$ &       1.0 \\
& $Compound \xrightarrow{\text{has mechanism}} Mechanism\ of\ Action \xrightarrow{\text{has assay target}} Gene \xrightarrow{\text{has link}} Gene$ &       0.8 \\
\bottomrule
\end{tabular}

\end{table}

\clearpage

\subsection{Explanations}

\begin{figure}[h!]
    \centering 
    \includegraphics[scale=0.18]{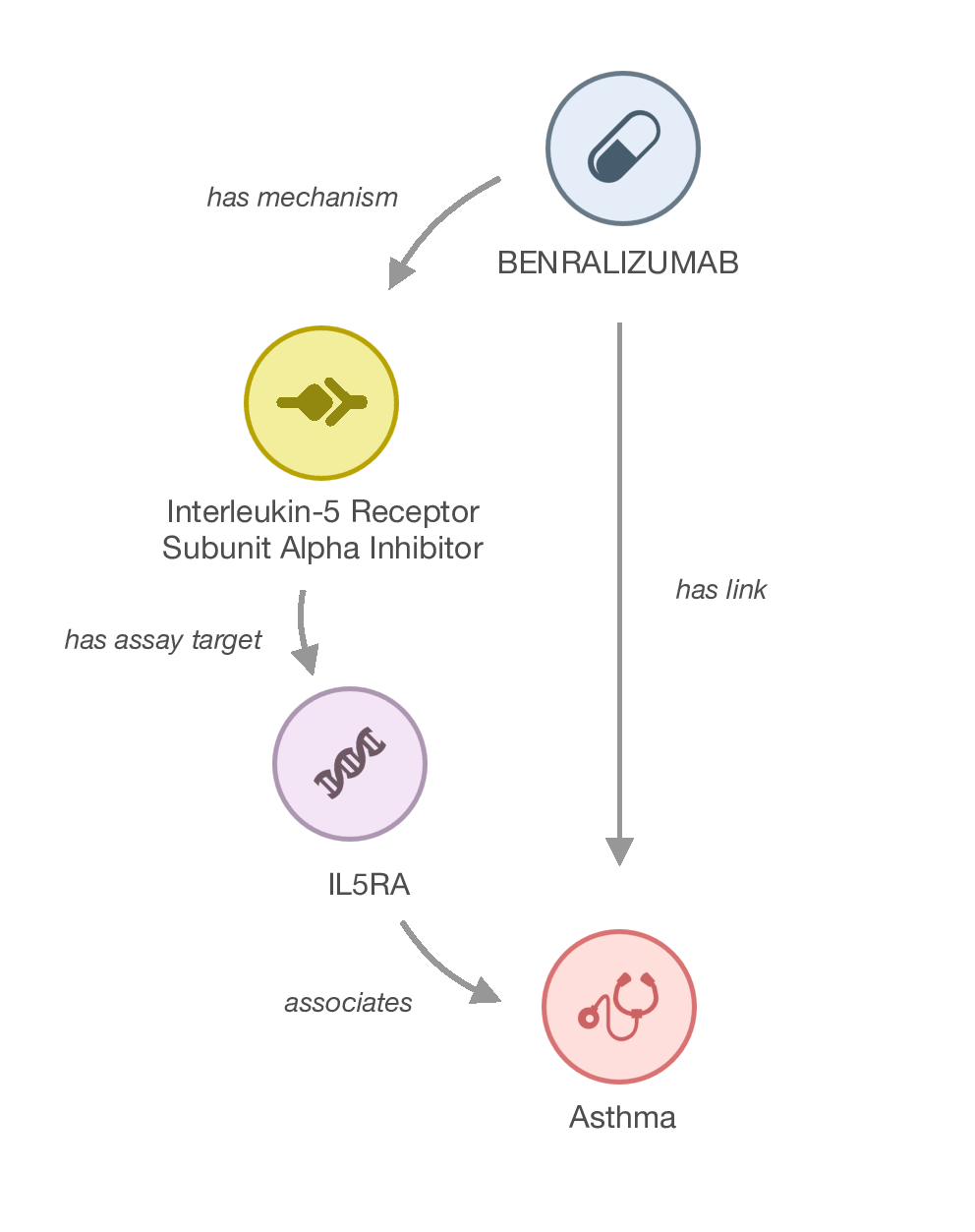}
    \caption{Explainable reasoning for recommending Benralizumab in the treatment of Asthma.}
    \label{fig:appendix_reasoning_path1}
\end{figure}

\begin{figure}[b!]
    \centering 
    \includegraphics[scale=0.18]{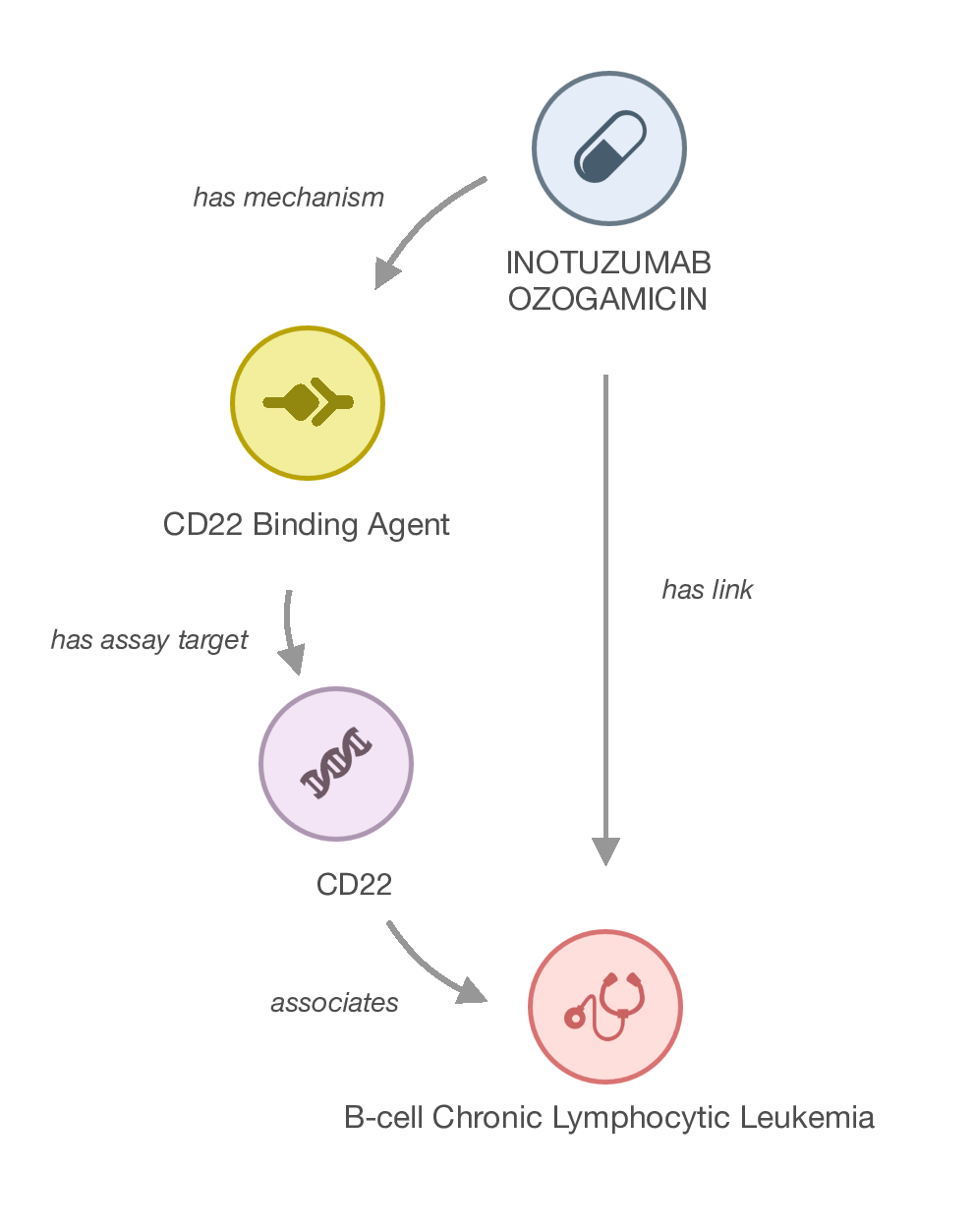}
    \caption{Explainable reasoning for recommending Inotuzumab ozogamicin in the treatment of B-cell Chronic Lymphocytic Leukemia.}
    \label{fig:appendix_reasoning_path2}
\end{figure}

\begin{figure}[b!]
    \centering 
    \includegraphics[scale=0.2]{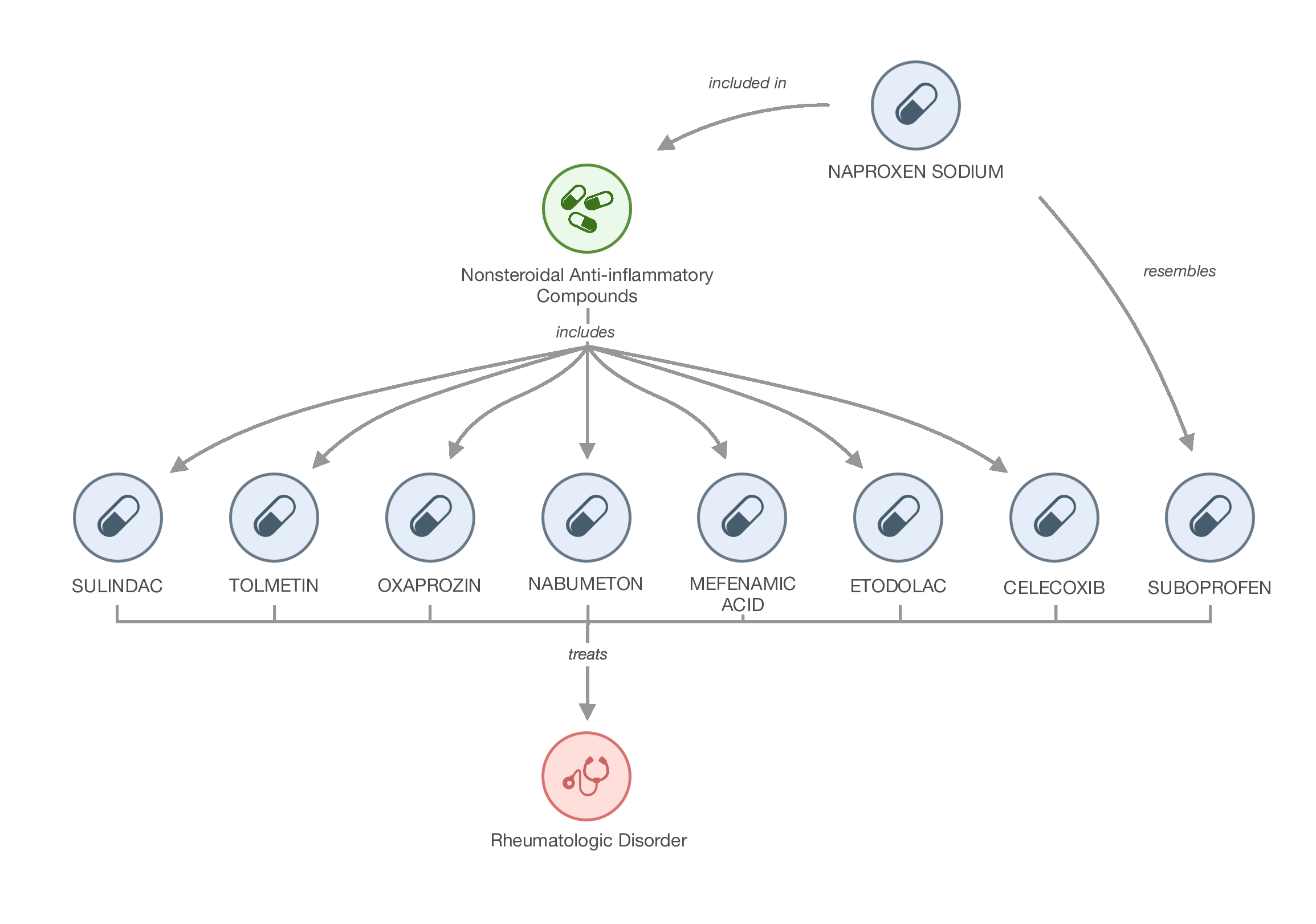}
    \caption{Explainable reasoning for recommending Naproxen Sodium in the treatment of rheumatic diseases.}
    \label{fig:appendix_reasoning_path3}
\end{figure}

\begin{figure}[b!]
    \centering 
    \includegraphics[scale=0.2]{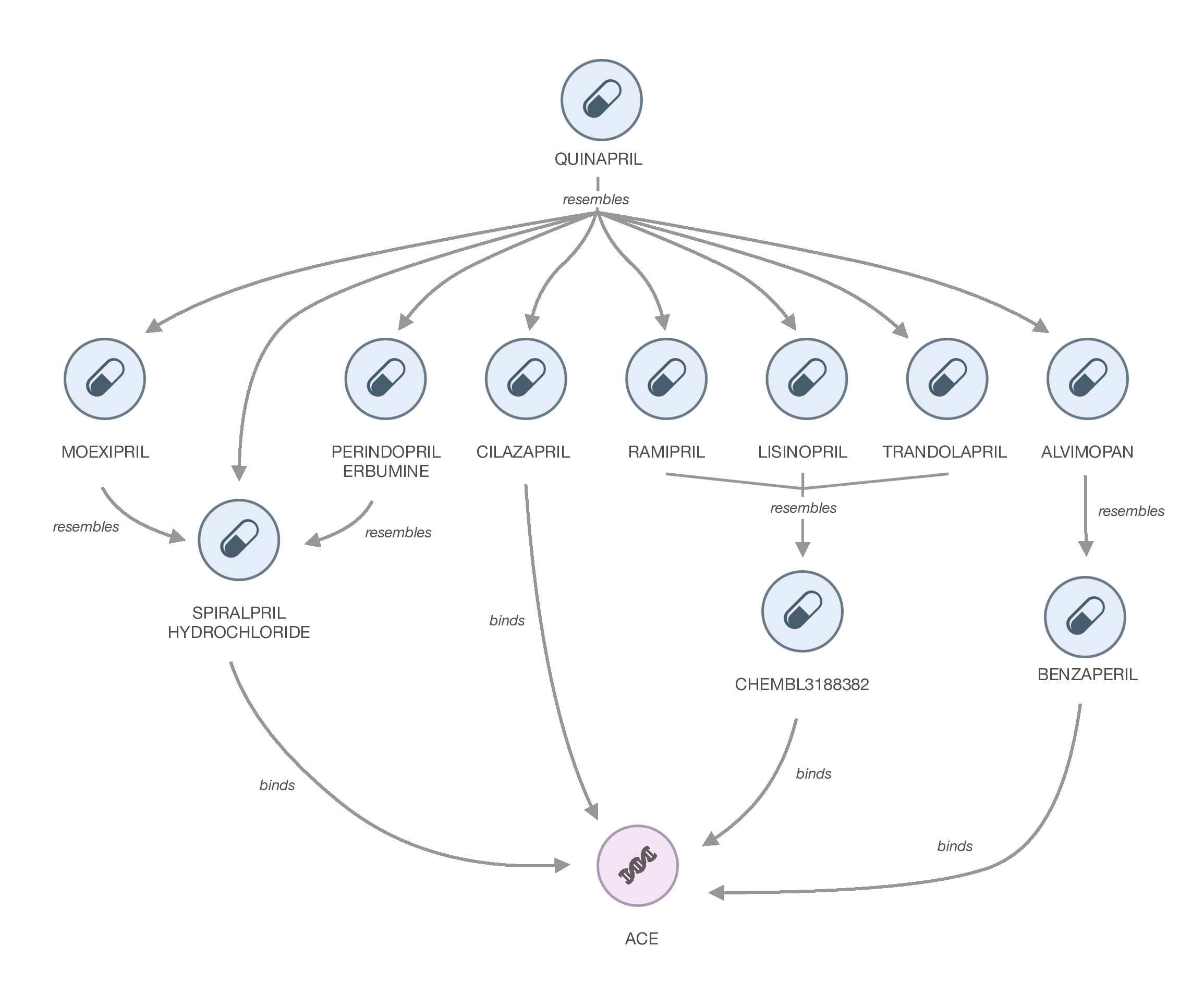}
    \caption{Explainable reasoning for recommending Quinapril as an ACE inhibitor.}
    \label{fig:appendix_reasoning_path4}
\end{figure}

\clearpage
\section{Pre-pruned Results}
\label{appendix:pre-pruned_results}

Below are the results before pruning of expected target node types was applied. This effectively keeps failed walks and predictions in the results sets.

\begin{table*}[h!]
\centering
\caption{Drug repurposing results (pre-pruning)}
{\footnotesize
\begin{tabular}{cccccc}

\textbf{Dataset}                        & \textbf{Model}   & \textbf{HITS@1} & \textbf{HITS@3} & \textbf{HITS@10} & \textbf{MRR} \\ \toprule
\multirow{6}{*}{\textbf{Hetionet}}       & TransE  & $.193\pm.027$      & $.364\pm.031$      & $.608\pm.034$       & $.312\pm.022$   \\[2pt]
                               & DistMult  & $.020\pm.014$      & $.086\pm.023$      & $.270\pm.022$       & $.096\pm.015$   \\[2pt]
                               & RGCN  & $.053\pm.027$      & $.200\pm.016$      & $.454\pm.036$       & $.217\pm.041$   \\[2pt]
                               & AnyBURL  & $.116\pm.013$      & $.288\pm.024$      & $.572\pm.036$       & $.258\pm.012$   \\[2pt]
                               & MINERVA & \cellcolor{blue!8}$\mathit{.281\pm.027}$      & \cellcolor{blue!8}$\mathit{.462\pm.055}$      & \cellcolor{blue!8}$\mathit{.701\pm.055}$       & \cellcolor{blue!8}$\mathit{.391\pm.034}$   \\[2pt]
                               & PoLo    & \cellcolor{green!25}$\mathbf{.306\pm.017}$      & \cellcolor{green!25}$\mathbf{.481\pm.029}$      & \cellcolor{green!25}$\mathbf{.742\pm.013}$       & \cellcolor{green!25}$\mathbf{.406\pm.015}$   \\ [2pt]\midrule
\multirow{6}{*}{\textbf{\begin{tabular}[c]{@{}c@{}}BIKG\\ Hetionet\end{tabular}}} & TransE  & $.016\pm.003$      & $.034\pm.005$      & $.088\pm.001$       & $.045\pm.004$   \\[2pt]
                               & DistMult  & $.012\pm.002$      & $.021\pm.003$      & $.046\pm.006$       & $.029\pm.003$   \\[2pt]
                               & RGCN  & $.056\pm.005$      & $.108\pm.008$      & \cellcolor{blue!8}$\mathit{.229\pm.011}$       & $.115\pm.004$   \\[2pt]
                               & AnyBURL  & $.000\pm.000$      & $.056\pm.005$      & $.205\pm.014$       & $.072\pm.002$   \\[2pt]
                               & MINERVA & \cellcolor{blue!8}$\mathit{.070\pm.007}$      & \cellcolor{blue!8}$\mathit{.137\pm.011}$      & $.204\pm.020$       & \cellcolor{blue!8}$\mathit{.116\pm.009}$   \\[2pt]
                               & PoLo    & \cellcolor{green!25}$\mathbf{.089\pm.005}$      & \cellcolor{green!25}$\mathbf{.162\pm.007}$      & \cellcolor{green!25}$\mathbf{.239\pm.008}$       & \cellcolor{green!25}$\mathbf{.142\pm.005}$   \\ [2pt]\midrule
\multirow{6}{*}{\textbf{\begin{tabular}[c]{@{}c@{}}BIKG\\ Hetionet+\end{tabular}}}          & TransE  & $.017\pm.002$      & $.052\pm.004$      & $.119\pm.011$       & $.057\pm.002$   \\[2pt]
                               & DistMult  & $.004\pm.002$      & $.013\pm.004$      & $.036\pm.009$       & $.025\pm.004$   \\[2pt]
                               & RGCN  & $.039\pm.008$      & $.091\pm.014$      & \cellcolor{blue!8}$\mathit{.191\pm.012}$       & \cellcolor{green!25}$\mathbf{.109\pm.060}$   \\[2pt]
                               & AnyBURL  & $.000\pm.000$      & $.065\pm.006$      & \cellcolor{green!25}$\mathbf{.205\pm.008}$       & $.070\pm.009$   \\[2pt]
                               & MINERVA & \cellcolor{green!25}$\mathbf{.069\pm.008}$      & \cellcolor{blue!8}$\mathit{.115\pm.010}$      & $.151\pm.011$       & \cellcolor{blue!8}$\mathit{.105\pm.009}$   \\[2pt]
                               & PoLo    & \cellcolor{blue!8}$\mathit{.068\pm.007}$      & \cellcolor{green!25}$\mathbf{.117\pm.011}$      & $.157\pm.011$       & \cellcolor{blue!8}$\mathit{.105\pm.008}$  \\[2pt]
\end{tabular}
}
\end{table*}

\begin{table*}[h!]
\centering
\caption{Drug repurposing results (pruned)}
{\footnotesize
\begin{tabular}{cccccc}

\textbf{Dataset}                        & \textbf{Model}   & \textbf{HITS@1} & \textbf{HITS@3} & \textbf{HITS@10} & \textbf{MRR} \\ \toprule
\multirow{6}{*}{\textbf{Hetionet}}       & TransE  & $.193\pm.027$      & $.364\pm.031$      & $.608\pm.034$       & $.312\pm.022$   \\[2pt]
                               & DistMult  & $.028\pm.002$      & $.138\pm.022$      & $.373\pm.039$       & $.129\pm.002$   \\[2pt]
                               & RGCN  & $.047\pm.028$      & $.194\pm.018$      & $.492\pm.043$       & $.221\pm.040$   \\[2pt]
                               & AnyBURL  & $.116\pm.013$      & $.288\pm.024$      & $.572\pm.036$       & $.258\pm.012$   \\[2pt]
                               & MINERVA & \cellcolor{blue!8}$\mathit{.351\pm.040}$      & \cellcolor{blue!8}$\mathit{.559\pm.058}$      & \cellcolor{blue!8}$\mathit{.790\pm.057}$       & \cellcolor{blue!8}$\mathit{.463\pm.041}$   \\[2pt]
                               & PoLo    & \cellcolor{green!25}$\mathbf{.378\pm.023}$      & \cellcolor{green!25}$\mathbf{.575\pm.025}$      & \cellcolor{green!25}$\mathbf{.845\pm.009}$       & \cellcolor{green!25}$\mathbf{.479\pm.015}$   \\ [2pt]\midrule
\multirow{6}{*}{\textbf{\begin{tabular}[c]{@{}c@{}}BIKG\\ Hetionet\end{tabular}}} & TransE  & $.016\pm.003$      & $.034\pm.005$      & $.088\pm.001$       & $.045\pm.004$   \\[2pt]
                               & DistMult  & $.004\pm.002$      & $.013\pm.002$      & $.036\pm.007$       & $.024\pm.003$   \\[2pt]
                               & RGCN  & $.056\pm.005$      & $.108\pm.008$      & $.239\pm.008$       & $.117\pm.005$   \\[2pt]
                               & AnyBURL  & $.000\pm.000$      & $.015\pm.001$      & $.046\pm.004$       & $.023\pm.001$   \\[2pt]
                               & MINERVA & \cellcolor{blue!8}$\mathit{.121\pm.002}$      & \cellcolor{blue!8}$\mathit{.194\pm.010}$      & \cellcolor{blue!8}$\mathit{.249\pm.013}$       & \cellcolor{blue!8}$\mathit{.160\pm.004}$   \\[2pt]
                               & PoLo    & \cellcolor{green!25}$\mathbf{.124\pm.005}$      & \cellcolor{green!25}$\mathbf{.204\pm.010}$      & \cellcolor{green!25}$\mathbf{.273\pm.016}$       & \cellcolor{green!25}$\mathbf{.167\pm.008}$   \\ [2pt]\midrule
\multirow{6}{*}{\textbf{\begin{tabular}[c]{@{}c@{}}BIKG\\ Hetionet+\end{tabular}}}          & TransE  & $.017\pm.002$      & $.052\pm.004$      & $.119\pm.011$       & $.057\pm.002$   \\[2pt]
                               & DistMult  & $.004\pm.002$      & $.014\pm.004$      & $.039\pm.001$       & $.026\pm.004$   \\[2pt]
                               & RGCN  & $.039\pm.008$      & $.104\pm.016$      & \cellcolor{green!25}$\mathbf{.222\pm.015}$       & \cellcolor{green!25}$\mathbf{.115\pm.005}$   \\[2pt]
                               & AnyBURL  & $.000\pm.000$      & $.056\pm.006$      & \cellcolor{blue!8}$\mathit{.205\pm.014}$       & $.072\pm.002$  \\[2pt]
                               & MINERVA & \cellcolor{green!25}$\mathbf{.076\pm.008}$      & \cellcolor{blue!8}$\mathit{.124\pm.011}$      & $.166\pm.013$       & \cellcolor{blue!8}$\mathit{.113\pm.010}$   \\[2pt]
                               & PoLo    & \cellcolor{blue!8}$\mathit{.074\pm.006}$      & \cellcolor{green!25}$\mathbf{.133\pm.010}$      & $.173\pm.012$       & \cellcolor{green!25}$\mathbf{.115\pm.007}$  \\[2pt]
\end{tabular}
}
\end{table*}

\begin{table*}[!htbp]
\centering
\caption{Drug-target interaction results (pre-pruning)}
{\footnotesize
\begin{tabular}{cccccc}

\textbf{Dataset}                        & \textbf{Model}   & \textbf{HITS@1} & \textbf{HITS@3} & \textbf{HITS@10} & \textbf{MRR} \\ \toprule
\multirow{6}{*}{\textbf{Hetionet}}       & TransE  & \cellcolor{green!25}$\mathbf{.287\pm.008}$      & \cellcolor{green!25}$\mathbf{.546\pm.013}$      & \cellcolor{green!25}$\mathbf{.918\pm.007}$       & \cellcolor{green!25}$\mathbf{.331\pm.005}$   \\[2pt]
                               & DistMult  & $.037\pm.004$      & $.120\pm.006$      & $.355\pm.010$       & $.111\pm.004$   \\[2pt]
                               & RGCN  & $.041\pm.017$      & $.123\pm.035$      & $.320\pm.087$       & $.104\pm.024$   \\[2pt]
                               & AnyBURL  & $.068\pm.018$      & $.128\pm.033$      & $.195\pm.049$       & $.210\pm.051$   \\[2pt]
                               & MINERVA & \cellcolor{blue!8}$\mathit{.159\pm.015}$      & \cellcolor{blue!8}$\mathit{.373\pm.037}$      & \cellcolor{blue!8}$\mathit{.756\pm.045}$       & \cellcolor{blue!8}$\mathit{.269\pm.021}$   \\[2pt]\midrule
\multirow{6}{*}{\textbf{\begin{tabular}[c]{@{}c@{}}BIKG\\ Hetionet\end{tabular}}} & TransE  & \cellcolor{green!25}$\mathbf{.193\pm.010}$      & \cellcolor{green!25}$\mathbf{.376\pm.012}$      & \cellcolor{green!25}$\mathbf{.674\pm.012}$       & \cellcolor{blue!8}$\mathit{.239\pm.004}$   \\[2pt]
                               & DistMult  & $.043\pm.007$      & $.106\pm.015$      & $.258\pm.016$       & $.095\pm.007$   \\[2pt]
                               & RGCN  & $.057\pm.025$      & $.142\pm.017$      & $.270\pm.028$       & $.123\pm.012$   \\[2pt]
                               & AnyBURL  & $.088\pm.003$      & $.167\pm.004$      & $.254\pm.007$       & \cellcolor{green!25}$\mathbf{.264\pm.004}$    \\[2pt]
                               & MINERVA & \cellcolor{blue!8}$\mathit{.104\pm.011}$      & \cellcolor{blue!8}$\mathit{.224\pm.014}$      & \cellcolor{blue!8}$\mathit{.589\pm.013}$       & $.172\pm.008$   \\[2pt]\midrule
\multirow{6}{*}{\textbf{\begin{tabular}[c]{@{}c@{}}BIKG\\ Hetionet+\end{tabular}}}          & TransE  & $.054\pm.004$      & $.098\pm.007$      & $.171\pm.011$       & $.080\pm.004$   \\[2pt]
                               & DistMult  & $.035\pm.006$      & $.083\pm.011$      & $.202\pm.028$       & $.083\pm.001$   \\[2pt]
                               & RGCN  & \cellcolor{blue!8}$\mathit{.058\pm.025}$      & \cellcolor{blue!8}$\mathit{.142\pm.017}$      & \cellcolor{green!25}$\mathbf{.270\pm.022}$       & \cellcolor{blue!8}$\mathit{.123\pm.012}$   \\[2pt]
                               & AnyBURL  &  \cellcolor{green!25}$\mathbf{.086\pm.002}$      & \cellcolor{green!25}$\mathbf{.162\pm.003}$      & $.251\pm.007$       & \cellcolor{green!25}$\mathbf{.263\pm.003}$   \\[2pt]
                               & MINERVA & $.026\pm.002$      & $.079\pm.006$      & \cellcolor{blue!8}$\mathit{.255\pm.027}$       & $.085\pm.004$   \\[2pt]
\end{tabular}
}
\end{table*}

\begin{table*}[!ht]
\centering
\caption{Drug-target interaction results (pruned)}
{\footnotesize
\begin{tabular}{cccccc}

\textbf{Dataset}                        & \textbf{Model}   & \textbf{HITS@1} & \textbf{HITS@3} & \textbf{HITS@10} & \textbf{MRR} \\ \toprule
\multirow{6}{*}{\textbf{Hetionet}}       & TransE  & \cellcolor{green!25}$\mathbf{.287\pm.008}$      & \cellcolor{green!25}$\mathbf{.546\pm.013}$      & \cellcolor{green!25}$\mathbf{.918\pm.007}$       & \cellcolor{green!25}$\mathbf{.331\pm.005}$   \\[2pt]
                               & DistMult  & $.039\pm.004$      & $.128\pm.007$      & $.367\pm.012$       & $.115\pm.005$   \\[2pt]
                               & RGCN  & $.060\pm.025$      & $.177\pm.057$      & $.376\pm.102$       & $.126\pm.031$   \\[2pt]
                               & AnyBURL  & $.169\pm.043$      & $.323\pm.081$      & $.493\pm.124$       & $.210\pm.051$   \\[2pt]
                               & MINERVA & \cellcolor{blue!8}$\mathit{.186\pm.015}$      & \cellcolor{blue!8}$\mathit{.420\pm.028}$      & \cellcolor{blue!8}$\mathit{.807\pm.039}$       & \cellcolor{blue!8}$\mathit{.296\pm.019}$   \\[2pt]\midrule
\multirow{6}{*}{\textbf{\begin{tabular}[c]{@{}c@{}}BIKG\\ Hetionet\end{tabular}}} & TransE  & $.193\pm.001$      & $.376\pm.012$      & \cellcolor{blue!8}$\mathit{.674\pm.012}$       & $.239\pm.004$   \\[2pt]
                               & DistMult  & $.044\pm.007$      & $.111\pm.014$      & $.269\pm.014$       & $.098\pm.007$   \\[2pt]
                               & RGCN  & $.173\pm.016$      & $.338\pm.013$      & $.539\pm.021$       & $.235\pm.008$   \\[2pt]
                               & AnyBURL  & \cellcolor{blue!8}$\mathit{.215\pm.005}$      & \cellcolor{blue!8}$\mathit{.408\pm.004}$      & $.623\pm.009$       & \cellcolor{blue!8}$\mathit{.264\pm.004}$   \\[2pt]
                               & MINERVA & \cellcolor{green!25}$\mathbf{.235\pm.005}$      & \cellcolor{green!25}$\mathbf{.516\pm.011}$      & \cellcolor{green!25}$\mathbf{.983\pm.015}$       & \cellcolor{green!25}$\mathbf{.305\pm.004}$   \\[2pt]\midrule
\multirow{6}{*}{\textbf{\begin{tabular}[c]{@{}c@{}}BIKG\\ Hetionet+\end{tabular}}}          & TransE  & $.054\pm.004$      & $.098\pm.007$      & $.171\pm.011$       & $.008\pm.004$   \\[2pt]
                               & DistMult  & $.035\pm.006$      & $.084\pm.011$      & $.206\pm.029$       & $.084\pm.001$   \\[2pt]
                               & RGCN  & \cellcolor{blue!8}$\mathit{.188\pm.008}$      & $.323\pm.019$      & $.539\pm.010$       & $.241\pm.006$   \\[2pt]
                               & AnyBURL  & \cellcolor{green!25}$\mathbf{.215\pm.003}$      & \cellcolor{green!25}$\mathbf{.408\pm.008}$      & \cellcolor{blue!8}$\mathit{.622\pm.005}$       & \cellcolor{green!25}$\mathbf{.263\pm.003}$   \\[2pt]
                               & MINERVA & $.181\pm.006$      & \cellcolor{blue!8}$\mathit{.368\pm.008}$      & \cellcolor{green!25}$\mathbf{.628\pm.019}$       & \cellcolor{blue!8}$\mathit{.243\pm.004}$   \\[2pt]
\end{tabular}
}
\end{table*}

\end{document}